\documentclass{article}

\PassOptionsToPackage{numbers, compress}{natbib}
\usepackage[preprint]{neurips_2025}
\usepackage{amsmath,amsfonts,bm}

\def\eqref#1{equation~\ref{#1}}
\def\1{\bm{1}}

\DeclareMathAlphabet{\mathsfit}{\encodingdefault}{\sfdefault}{m}{sl}
\SetMathAlphabet{\mathsfit}{bold}{\encodingdefault}{\sfdefault}{bx}{n}

\title{MLFM: Multi-Layered Feature Maps for Richer Language Understanding in Zero-Shot Semantic Navigation}

\author{
Sonia Raychaudhuri$^1$,  Enrico Cancelli$^2$, Tommaso Campari$^{3}$,\\ \textbf{Lamberto Ballan}$^2$,
\textbf{Manolis Savva}$^1$, \textbf{Angel X. Chang}$^{1,4}$ \\
$^1$Simon Fraser University, $^2$University of Padova, \\
$^3$Fondazione Bruno Kessler (FBK),
$^4$Alberta Machine Intelligence Institute (Amii)\\
\small{\url{https://github.com/3dlg-hcvc/langmonmap}}
}

\usepackage{hyperref}
\usepackage{url}

\usepackage{graphicx}
\usepackage{amsmath}
\usepackage{amssymb}
\usepackage{amsthm}

\usepackage{booktabs}
\usepackage{algorithm}
\usepackage{algorithmic}

\usepackage[capitalize]{cleveref}
\crefname{section}{Sec.}{Secs.}
\Crefname{section}{Section}{Sections}
\Crefname{table}{Table}{Tables}
\crefname{table}{Tab.}{Tabs.}

\usepackage{times}

\usepackage{tikz}
\usepackage{comment}
\usepackage{amsmath,amssymb} % define this before the line numbering.
\usepackage{color}
\usepackage{mdframed}
\usepackage{custom_symbols_iclr}

\usepackage{float}
\usepackage[accsupp]{axessibility}  % Improves PDF readability for those with disabilities.

\usepackage{multirow}
\usepackage{colortbl}
\usepackage{hhline}
\usepackage{paralist}
\usepackage{makecell}
\usepackage{xspace}
\usepackage{wrapfig}
\usepackage{listings}

\definecolor{linkcolor}{HTML}{991408}  % red
\definecolor{citecolor}{HTML}{2E7E2A}  % green
\definecolor{filecolor}{HTML}{131877}  % dark blue
\definecolor{menucolor}{HTML}{727500}  % yellow
\definecolor{runcolor} {HTML}{137776}  % teal
\definecolor{urlcolor} {HTML}{0a2bbf}  % blue
\hypersetup{colorlinks=true,linkcolor=linkcolor,citecolor=citecolor,filecolor=filecolor,menucolor=menucolor,runcolor=runcolor,urlcolor=urlcolor}

\begin{document}
\maketitle
\begin{abstract}
Recent progress in large vision-language models has driven improvements in language-based semantic navigation, where an embodied agent must reach a target object described in natural language. Yet we still lack a clear, language-focused evaluation framework to test how well agents ground the words in their instructions.
We address this gap by proposing \emph{\DATASET}, an open-vocabulary multi-object navigation dataset with natural language goal descriptions (e.g. `\emph{go to the red short candle on the table}') and corresponding fine-grained linguistic annotations (e.g., attributes: color=red, size=short; relations: support=on).
These labels enable systematic evaluation of language understanding. To evaluate on this setting, we extend multi-object navigation task setting to \emph{\task} (\emph{\taskshort}), where the agent must find a sequence of goals specified using language.
% specifically created to test an agent's ability to locate objects described at different levels of detail, from broad category names to fine attributes and object–object relations. 
% Every description in \DATASET was manually checked, yielding a lower error rate than existing lifelong- and semantic-navigation datasets.
% On top of \DATASET we build \benchmarkname, a benchmark that measures how well current semantic-navigation methods understand and act on these descriptions while moving toward their targets. \benchmarkname allows us to systematically compare models on their handling of attributes, spatial and relational cues, and category hierarchies, offering the first thorough, language-centric evaluation of embodied navigation systems.
Furthermore, we propose \emph{\methodname} (\emph{\methodnameshort}), a novel method that builds a queryable, multi-layered semantic map from pretrained vision-language features and proves effective for reasoning over fine-grained attributes and spatial relations in goal descriptions.
% , particularly effective when dealing with small objects or instructions involving spatial relations.
Experiments on \emph{\DATASET} show that \emph{\methodnameshort} outperforms state-of-the-art zero-shot mapping-based navigation baselines.
% Across \DATASET and GOAT-bench, \methodnameshort consistently outperforms state-of-the-art mapping-based navigation baselines.
\end{abstract}
\section{Introduction}
\label{sec:intro}
\begin{figure}[t]
  \centering
  \includegraphics[trim={0 4cm 0 0},width=\linewidth]{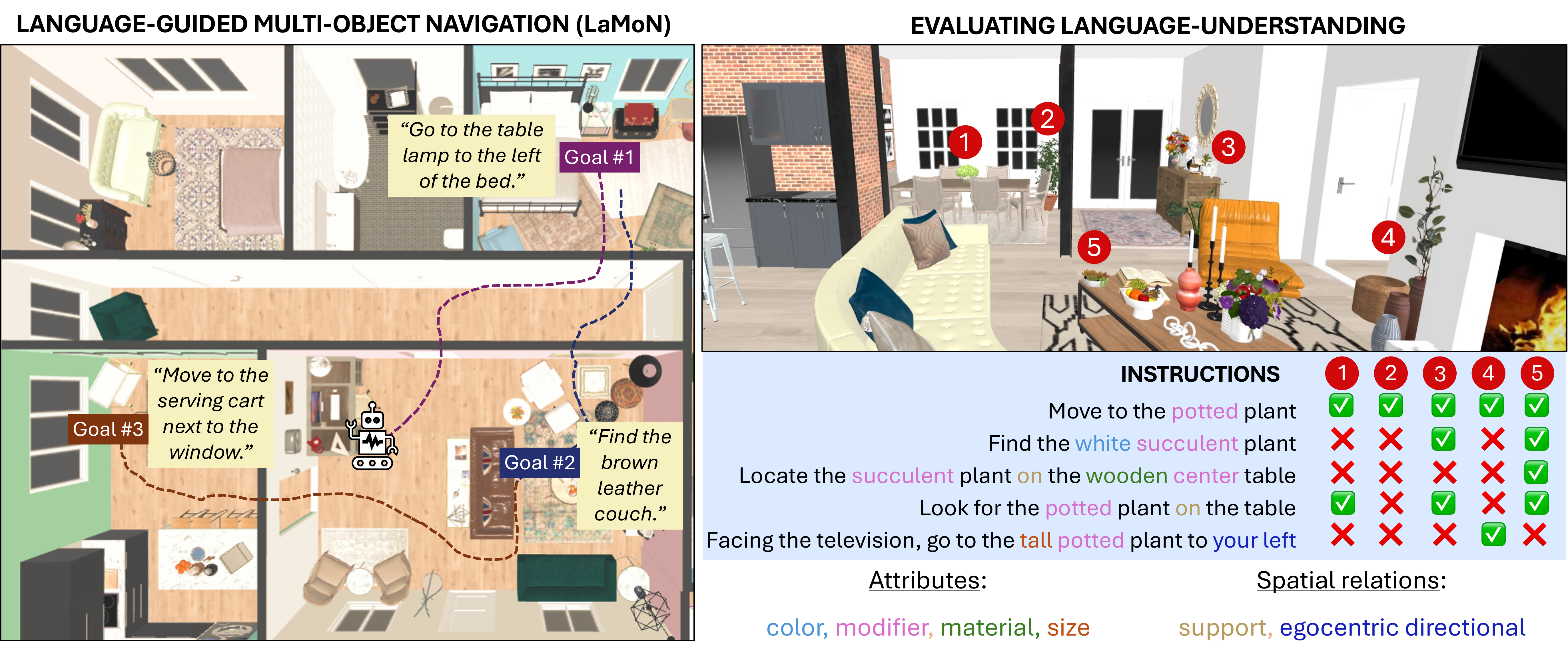}
  \caption{
  \textbf{\task (\taskshort)} requires an agent to navigate to multiple goals, described using descriptions (left).
  We evaluate fine-grained language understanding by tagging each description with attributes (\emph{color}) and spatial relations (\emph{support}) (right). There may be multiple positive matches (objects matching all attributes and relations in the instruction) and the agent is scored correct if it stops at any (\textcolor{green}{check} marks a match and \textcolor{red}{cross} marks a non-match). 
  % \textbf{\benchmarkname} provides a means to evaluate natural language understanding in semantic navigation methods by introducing \DATASET dataset that not only uses natural language to describe goal objects but also stores linguistic annotations for each description. (Left) a single object can be described using descriptions with varying specificity ranging from no attribute to using multiple attributes. (Right) semantic navigation gets harder with increasing number of attributes in the description, since it requires the agent to navigate to a very specific object instance.
  }
  \label{fig:teaser}
\vspace{-12pt}
\end{figure}

% \tc{
Semantic navigation is a rapidly growing sub-field in embodied AI~\citep{deitke2022retrospectives} where an agent is tasked with navigating to a target object described using either an object category~\citep{anderson2018evaluation,batra2020objectnav,yokoyama2024hm3d}, image~\citep{krantz2022instance} or natural language descriptions~\citep{li2021ion,taioli2024collaborative}. Each of these target specifications poses unique challenges. Among them, natural language is the most natural way for humans to express goals, yet it is also inherently ambiguous and context-dependent. 
In this paper, we focus on natural language guided semantic navigation, where goals are specified using language descriptions, for example, `\emph{Go to the white table lamp on the wooden corner table}' (\Cref{fig:teaser}). 

% To construct datasets with descriptions, 
Prior efforts to build datasets with language descriptions~\citep{khanna2024goat} often rely on large vision-language models (VLMs) to extract attributes from images and large language models (LLMs) to merge them into sentences. 
However, VLMs frequently hallucinate attributes~\citep{li2023evaluating}, especially in cluttered or out-of-distribution scenes, introducing noise that biases both agent learning and evaluation. 
% Such hallucinated attributes introduce significant noise, biasing both agent learning and fair evaluation. 
To mitigate this, we propose the \emph{\DATASET} dataset derived from the Habitat Synthetic Scenes Dataset (HSSD)~\citep{khanna2023hssd}. \DATASET uses ground-truth attributes for objects, with all descriptions manually validated to avoid VLM errors. Furthermore, it provides fine-grained linguistic annotations--capturing object attributes (e.g. color, size, material) and spatial relations (e.g. `on', `near', `left')--to enable systematic evaluation of language understanding.

Vision-and-language navigation (VLN)~\citep{anderson2018vision,ku2020room,raychaudhuri2025zeroshotobjectcentricinstructionfollowing} differs from our setting in that it provides full route instructions (e.g., ``\emph{leave the bedroom, turn left, pass the painting, and stop at the sofa}''), which are mostly crowdsourced and rarely contain hallucinated attributes. By contrast, object navigation specifies only the goal object, coupling grounding with exploration. In real-world interaction, people rarely provide step-by-step paths; instead, they say, ``\emph{bring me the blue mug on the bedside table}''.
% Prior work in vision-and-language navigation (VLN) instructs the agent with a \emph{full route}—for example, “leave the bedroom, turn left, pass the painting, and stop at the sofa” \citep{anderson2018vision,ku2020room,raychaudhuri2025zeroshotobjectcentricinstructionfollowing}.
% Because these descriptions are crowdsourced, they rarely contain hallucinated attributes.  However, VLN differs from object navigation tasks that specifies just target object description as the VLN task couples object grounding with the added challenge of parsing route directives.
% In everyday settings, people rarely spell out an entire path; they simply name the goal—“bring me the blue mug on the bedside table.”
Goal-only language is therefore both \emph{natural} and \emph{challenging}: the agent must explore unseen space, decide when it has found the correct object, and remember what it has already observed.
Sequential tasks such as Multi-Object Navigation (MultiON) \citep{wani2020multion,Raychaudhuri_2024_WACV} and GOAT-Bench~\citep{chang2023goat,khanna2024goat} capture this difficulty by revealing each target only after the previous one is reached.  
% Prior benchmarks on this task setting has so far offered limited linguistic variety.
We extend such task settings by proposing \emph{\task} (\emph{\taskshort}) where each of three sequential goals is specified through a natural language description with different levels of specificity.
For instance, in \Cref{fig:teaser}, depending on the level of specificity, the description can match just one or multiple target objects.
% (\Cref{fig:teaser}). 
% Each episode presents three goals in succession, each described in natural language that may include colour, size, material, or support relations. 
Success in \taskshort\ requires two key abilities: ($i$) building a semantic map memory while exploring, and ($ii$) querying that memory to reason about objects whose descriptions vary in granularity. To this end, we propose a novel method \emph{\methodname} (\emph{\methodnameshort}) that constructs a multi-layered semantic map, leverages it to identify the goal object, semantically explores the environment and navigates via a path planner in an unseen environment.
% \benchmarkname focuses on goal-specific language instructions and offers a testbed for systematically evaluating these two aspects in map-based navigation methods.
% By focusing on goal-only instructions, \benchmarkname isolates these skills and offers a principled testbed for evaluating map-based navigation methods.

In summary, our contributions are threefold:
% (i) \emph{\benchmarkname}: a benchmark to evaluate semantic navigation agents on their ability to understand natural language based on various linguistic cues (color, size, texture, state, number, material, lexical modifiers, and support relations); 
(i) \emph{\DATASET}: an open-vocabulary multi-object navigation dataset with natural language goal descriptions and fine-grained linguistic annotations, enabling systematic evaluation of language understanding;
(ii) \emph{\taskshort}: an extension of MultiON, where  goals are specified using language descriptions at varying levels of specificity;
% with varying amount of specificity and language cues that exist in the description;
(iii) \emph{\methodnameshort}: a multi-layer semantic-mapping approach that preserves rich object features 
% without the cost of full 3D mapping 
and achieves improved performance over state-of-the-art zero-shot map-based semantic navigation baselines on the \DATASET dataset.
\section{Related Work}
\label{sec:related}

\mypara{Semantic navigation.}
In embodied AI, one of the basic semantic navigation tasks is the Object Navigation (ObjectNav), where an agent is required to navigate to an object specified by its category. While earlier benchmarks~\citep{anderson2018evaluation,batra2020objectnav} have used a closed set of object categories, HM3D-OVON~\citep{yokoyama2024hm3d} introduces a benchmark containing open-vocabulary categories to describe the goal.
Multi-Object Navigation (MultiON)~\citep{wani2020multion} extends this single object navigation to a multi-object setting where multiple objects are made available to the agent in sequence. MultiON requires the agent to remember objects observed in the past so as to come back if needed, for efficiency. Another type of semantic navigation is Instance-Nav~\citep{krantz2022instance}, where the agent is given an image of the goal object. While this task has its own challenges with respect to finding a specific object instance observed from a specific view angle, it does not reflect real-world scenarios where it is more likely for someone to describe an object of interest in words.
% \mypara{Language-based semantic navigation.}
A more natural semantic navigation is where the goal object is described using natural language, a task called language-based InstanceNav~\citep{li2021ion,taioli2024collaborative}.
GOAT-Bench~\citep{khanna2024goat} was introduced to provide a benchmark on multi-object multi-modal navigation, combining different modalities (object category, image and language) to specify the goals.
% and the agent is tasked with finding multiple goals in sequence.
Compared to GOAT-Bench, where the language descriptions are generated via a VLM and contain significant errors, our \DATASET dataset contains manually validated language descriptions at varying levels of specificity as well as annotations for fine-grained linguistic tags.
% In this paper we focus on a single modality, language, to specify goals in a multi-object navigation task, extending on GOAT-Bench.

% Recent developments of Object-Navigation benchmarks have shifted the focus of navigation from object classes to specific object instances corresponding to a set of individual characteristics. 
% Such goal instances can be described using either an image~\citep{krantz2022instance} or a language description~\citep{li2021ion,taioli2024collaborative}. In language-based ObjectNav, the description often contains various linguistic features such as color,material, size, texture, etc. as well as spatial relations between the goal and other objects.
% GOAT-Bench~\citep{khanna2024goat} has recently introduced a benchmark that focuses on multimodal lifelong navigation, to evaluate how a single method can navigate to goals specified using both image and language descriptions (using either a coarse category or a specific object description).
% Our proposed model and task focus on exploring Multi-Object-Nav with only language-based goal descriptions and aims to gauge performance against various types of instruction with different linguistic features, varying levels of detail, and broad spectrum of object variations. Moreover, compared to GOAT-Bench which uses a Vision-Language-Model (VLN) like BLIP-2~\citep{li2023blip} to extract object attributes, we use ground-truth semantic information that already exists in the synthetic HSSD scene dataset, making our proposed benchmark cleaner and free of errors.

\mypara{Semantic maps.}
A widely used representation in semantic navigation is a semantic map that serves as a memory of observations that the agent has made throughout its navigation. It has been shown to improve efficiency in the longer-horizon MultiON task~\citep{khanna2024goat} by remembering objects of importance. While some works~\citep{chaplot2020object,gervet2023navigating,zhang2025mapnav} store explicit information such as object category in the map, more recent methods store implicit features in the map~\citep{huang2023visual,gadre2023cows,chen2023object}, enabling open-vocabulary memory representation. These features are often extracted from a large VLM such as CLIP~\citep{radford2021learning}. 
VLMaps~\citep{huang2023visual} uses pixel-level embeddings from LSeg~\citep{li2022language}, while VLFM~\citep{yokoyama2023vlfm} uses BLIP-2~\citep{li2023blip} image features to build a 2D top-down map and compute semantic frontiers to explore the environment while performing navigation. Others build 3D maps~\citep{zhang20233d,conceptgraphs} which preserve geometric details of the objects and allow better reasoning, but are very costly to build and maintain, especially in long-horizon tasks.
In this work, we propose a \emph{multi-layer map} that serves as a middle-ground between a 2D and a 3D grid map and contains stacked layers of 2D maps, providing more granularity and flexibility than a 2D map but avoids the cost of a full 3D grid map.

\mypara{Evaluation benchmarks and datasets.}
% With the tremendous progress in large language models (LLMs) and large vision-language models (VLMs), 
Current semantic navigation methods focus on task success, i.e. whether the agent is able to find a goal, but lack an evaluation framework to test language understanding. At the same time, in other fields such as 3D scene understanding and grounding, evaluating AI models against natural language understanding has gained a lot of interest. OpenLex3D~\citep{kassab2025openlex3d} proposes a benchmark that evaluates open-vocabulary scene representation methods by introducing new label annotations, such as synonyms, depictions, visual similarity, and clutter, in existing scene datasets. 
Eval3D~\citep{duggal2025eval3d} proposes an evaluation tool to evaluate 3D generative models to assess their geometric and semantic consistencies.
Agentbench~\citep{liu2023agentbench} introduces a testbed to systematically evaluate LLMs acting as agents on their reasoning capabilities. 
ViGiL3D~\citep{wang2024vigil3d} introduces a dataset for evaluating visual grounding methods with respect to a diverse
set of language patterns, such as object attributes, relationships between objects, target references, etc. 
Motivated by these works, we introduce \emph{\DATASET}, a dataset containing fine-grained linguistic annotations to evaluate semantic navigation methods. 
% with respect to object attributes and relationships. 
% extracted from the language descriptions that specify goal objects in our dataset.
\section{Task and Dataset}
\label{sec:task}
\mypara{Task.}
Our \emph{\task}~(\emph{\taskshort}) task is an extension of the MultiON task~\citep{wani2020multion,Raychaudhuri_2024_WACV} and the GOAT-Bench task~\citep{chang2023goat,khanna2024goat}, where an agent is spawned in an unseen indoor environment and tasked with sequentially navigating to multiple goal objects. 
While MultiON uses a closed-vocabulary category name to describe each goal, GOAT-Bench uses multiple modalities--image, category and open-vocabulary language descriptions--to describe each goal. In contrast, we set up our task in an open-vocabulary setting, specifying each goal using \emph{language descriptions with different levels of specificity}.
Each episode contains \emph{three} goals, disclosed to the agent one at a time. 
% rather than all at once. 
% Revealing every target at the start would let the agent approach each object the moment it first appears, undermining our aim of testing how well it stores and reuses information gathered during exploration. 
Language descriptions in \emph{\taskshort} describe objects with varying levels of specificity, e.g. `go to the couch' vs. `go to the black couch' vs. `go to the black three piece L-shaped sectional couch'.
This allows us to evaluate the agent's capability to understand coarse vs fine-grained linguistic cues in the descriptions.
% to identify objects specified without any attribute as well as those specified by one or more attributes.
At every step, the agent takes as inputs egocentric RGB-D images, GPS and compass readings relative to the agent's starting pose in the episode, and the language instruction for the current goal $g_i$, The agent takes one of four actions: \textit{move forward} by 25 cm, \textit{turn left} by 30° or \textit{turn right} by 30°, and \textit{found}. A goal is successful if the agent generates \textit{found} within 1.5 m of the goal object and under 500 steps. The episode continues even when the agent fails to navigate to one goal.

\subsection{Dataset}
\label{sec:dataset}
In  this section, we introduce the \emph{\DATASET} dataset that implements the \taskshort~task.
Among the prior datasets to evaluate various semantic navigation tasks, GOAT-Bench~\citep{khanna2024goat} is the closest to ours (\Cref{tab:dataset_comparison}). However, due to the use of a VLM during creation without manual verification, their language goals contain significant errors (see \Cref{sec:dataset_more}).

\begin{table}
\vspace{-10pt}
\caption{\textbf{Comparison with prior datasets.} 
\DATASET contains open-vocabulary object+instance goals, annotated with fine-grained linguistic tags and language descriptions are manually verified and exist at different levels of specificity (\emph{Specificity}), depending on which it can match with one or multiple correct target objects.
% \DATASET provides ObjectNav as well as language-based InstanceNav goals, with additional linguistic tags for the goal descriptions.
}
\label{tab:dataset_comparison}
\centering
\resizebox{\linewidth}{!}{
\begin{tabular}{llcccccc}
\toprule
Datasets & Task &\thead{ObjectNav\\goals} & \thead{InstanceNav\\language goals} & \thead{Open-\\vocab} & \thead{Linguistic\\tags} &\thead{No significant \\description\\errors} &\thead{Specif-\\icity} \\
\midrule
\rowcolor{gainsboro}
OVMM\citep{yenamandra2023homerobot} &Manipulation &\textcolor{tblgreen}{$\checkmark$}  &\textcolor{red}{$\times$} &\textcolor{tblgreen}{$\checkmark$}  &\textcolor{red}{$\times$} &- &- \\
\midrule
ObjectNav\citep{batra2020objectnav,habitatchallenge2022} &Object nav &\textcolor{red}{$\times$} &\textcolor{red}{$\times$} &\textcolor{red}{$\times$} &\textcolor{red}{$\times$}  &- &- \\
HM3D-OVON\citep{yokoyama2024hm3d} &Object nav &\textcolor{tblgreen}{$\checkmark$}  &\textcolor{red}{$\times$} &\textcolor{tblgreen}{$\checkmark$} &\textcolor{red}{$\times$}  &-  &-\\
MultiON\citep{wani2020multion,Raychaudhuri_2024_WACV} &MultiON  &\textcolor{tblgreen}{$\checkmark$}  &\textcolor{red}{$\times$} &\textcolor{red}{$\times$} &\textcolor{red}{$\times$}  &-  &-\\
OneMap\citep{busch2024mapallrealtimeopenvocabulary} &Object nav, MultiON &\textcolor{tblgreen}{$\checkmark$}  &\textcolor{red}{$\times$} &\textcolor{red}{$\times$} &\textcolor{red}{$\times$} &-  &- \\
Goat-Bench\citep{khanna2024goat} &Multi-(object+instance) nav   &\textcolor{tblgreen}{$\checkmark$}  &\textcolor{tblgreen}{$\checkmark$}  &\textcolor{tblgreen}{$\checkmark$}   &\textcolor{red}{$\times$} &\textcolor{red}{$\times$} &\textcolor{red}{$\times$}  \\
\midrule
\emph{\DATASET} &Multi-(object+instance) nav   &\textcolor{tblgreen}{$\checkmark$}  &\textcolor{tblgreen}{$\checkmark$}  &\textcolor{tblgreen}{$\checkmark$}   &\textcolor{tblgreen}{$\checkmark$}  &\textcolor{tblgreen}{$\checkmark$} &\textcolor{tblgreen}{$\checkmark$}  \\
\bottomrule
\end{tabular}}
\vspace{-8pt}
\end{table}

\mypara{Episode generation.}
We use high-quality synthetic 3D scenes from HSSD~\citep{khanna2023hssd}.
For each episode we draw a random navigable start pose for the agent and choose three goals from the objects in the scene. 
For each goal, we store the \emph{language description}, \emph{linguistic tags} along with the \emph{viewpoints} or navigable positions around the goal (see \Cref{sec:dataset_more} for detail).

\mypara{\newadd{Attribute descriptions.}}
We convert the ground-truth attributes available in HSSD into fluent object descriptions with GPT-4~\citep{openai2023gpt4}.  
The model is prompted with in-context examples~\citep{brown2020language,liu2021makes,wu2023self} to form a coherent natural language description.
% \emph{``You are given a list of attributes for an object. Using them, form a coherent description such that a human can find the object in a scene"}.
% Each description is then wrapped in a standard navigation template (e.g., \texttt{Find …}, \texttt{Go to …}, etc.) to create the final instruction.  
% All sentences are manually reviewed and, where needed, edited to correct residual LLM errors.  
Next, we use GPT-4 to also tag every instruction with the linguistic cues~\citep{wang2024vigil3d} and release the annotations along with the dataset:  \begin{inparaitem}
\item \textbf{Color}—a color adjective in the description, e.g.\ `the \textit{brown} chair'.
\item \textbf{Size}—an explicit size cue, e.g.\ `the \textit{small} candle'.
\item \textbf{Texture}—a surface pattern or finish, e.g.\ `\textit{knitted} bath mat'.
\item \textbf{State}—a transient state that could change, e.g.\ `\textit{illuminated} makeup mirror'.
\item \textbf{Number}—a numeral in the description, e.g.\ `the cabinet with \textit{two} doors'.
\item \textbf{Material}—the substance an object is made of, e.g.\ `\textit{wooden} shoe rack'.
\item \textbf{Modifier}—any other adjectival qualifier, e.g.\ `the \textit{easy} chair'.
% \item \textbf{Support relationship}—an explicit support relation, e.g.\ `the potted plant \textit{on} the console table'.
\end{inparaitem}

\mypara{\newadd{Spatial relation descriptions.}}
\newadd{We programmatically extract spatial relationships between a pair of objects from the ground-truth 3D bounding boxes in HSSD, following methodologies similar to prior works in text-conditioned 3D scene generation~\citep{chang2014learning,tam2025sceneeval} (see \Cref{sec:dataset_more}).}
We extract the following spatial relations: 
\begin{inparaitem}
\item \textbf{Egocentric directional}--a cardinal direction relative to the agent's orientation, e.g.\ `the plant to \emph{your left}'.
\item \textbf{Allocentric}--a spatial relation relative to other objects independent to the agent's current pose and further sub-divided into:
\begin{inparaenum}[(a)]
    \item \textbf{Support}--an explicit support relation, e.g.\ `the potted plant \textit{on} the console table'.
    \item \textbf{Directional}--a direction-dependent spatial arrangement which is not a support relation, e.g.\ `the mirror \emph{above} the sink'.
    \item \textbf{Proximity}--a relation encoding the notion of metric or qualitative closeness, e.g.\ `the floor lamp \emph{near} the bed' or `the plant \emph{next to} the couch'.
    \item \textbf{Containment}--an inclusion relationship, e.g.\ `the plush toy \emph{inside} the crib'.
\end{inparaenum}
\end{inparaitem}

\mypara{Statistics.}
We generate two splits for our dataset: validation and test, with distinct set of target objects and distinct scenes. The validation split contains \newadd{932} episodes, each with three goals, totaling \newadd{2796} goal descriptions spanning 20 scenes. The test split contains \newadd{855} episodes, each with three goals, totaling \newadd{2565} goal descriptions spanning 15 scenes (see \Cref{sec:dataset_more}). 

\section{Method}
\label{sec:method}

% We introduce \emph{\methodnameshort} that iteratively builds a multi-layer feature map, using which it semantically explores the environment and then navigates towards the goal using a path planner. We use an open-vocabulary object detector in addition to the multi-layer map to confirm that a goal has been found, an approach known as consensus filtering~\citep{busch2024mapallrealtimeopenvocabulary}. We then use an A* path planner to navigate to a goal when found.
We propose \emph{\methodname} (\emph{\methodnameshort}), a zero-shot semantic navigation method that incrementally builds a semantic map by storing visual features from a pretrained vision-language model~\citep{radford2021learning}. At each step, \methodnameshort queries the map by comparing stored features with the goal's language embedding, generating a similarity heatmap over cells. If a likely target cell is found, the agent navigates to it via a path planner; otherwise, it continues exploring using a semantic exploration method VLFM~\citep{yokoyama2023vlfm}.
% For composite queries with spatial relations (e.g., “candle on the night-stand”) the kernel contains separate embeddings for the supported and supporting objects and is evaluated across height bands. 
% When a map cell is marked as a possible target, the agent moves toward it using a path planner.
% % an open-vocabulary detector checks the prediction, and the goal is accepted only if both sources agree i.e., via consensus filtering~\citep{busch2024mapallrealtimeopenvocabulary}. 
% % An A* planner is used to compute the agent's actions.
% In case the goal has not been observed yet, the agent keeps exploring using VLFM~\citep{yokoyama2023vlfm}.

% \subsection{Navigation pipeline}
\mypara{Two phase navigation (EAE-E).}
\label{sec:phases}
For navigation, we introduce a two-phase method called \emph{EAE-E}.
At the beginning of an episode the map is still sparse and because it stores \emph{features} rather than explicit class labels, its cosine scores can be noisy.  
An open-vocabulary detector, e.g. YOLO-World~\citep{cheng2024yolo}, on the other hand, yields high-precision category predictions whenever the object is in view.  
Hence, early on the detector is the more reliable signal while the map mainly proposes candidate locations that must later be verified by the detector. As exploration continues and the map densifies, trust gradually shifts toward the map.
Following this idea we introduce two-phase navigation:  
($i$)~\textbf{Explore-And-Exploit (EAE).} For the first $\mathcal{N}_{e}$ steps a location is accepted as the goal only if \emph{both} the similarity map $\mathcal{S}$ 
% \axc{$\mathcal{M}$ was defined as the feature map} 
and the detector agree, similar to consensus filtering~\citep{busch2024mapallrealtimeopenvocabulary}. Upon agreement the agent moves to that cell via the path planner; otherwise it resumes exploration.  
($ii$)~\textbf{Exploit-only (E).} For the remaining ($\mathcal{N}-\mathcal{N}_{e}$) steps it relies solely on the map, selecting the cell with highest similarity to thes goal description and navigates to it.

\subsection{Mapping} 
% \mypara{Map building.}
% In this section we discuss the map building method in \methodnameshort.
% \input{figures/fig-method-iclr}
The motivation behind using a multi-layered map is that it allows us to capture the vertical context without incurring the cubic memory cost of a full 3D voxel grid.
Though flattening everything into a single 2D plane is a commonly used approach, doing so merges features from stacked objects—small items vanish beneath larger ones—and discards the height cues needed for reasoning about spatial relations between objects.
We therefore adopt a \emph{multi-layer} top-down map that stores only a handful of horizontal slices: granular enough to separate objects at different heights, yet linear in memory and update cost.
\begin{figure}[t]
  \centering
  % \vspace{-9pt}
  \includegraphics[trim={0 2cm 0 1cm},width=\linewidth]{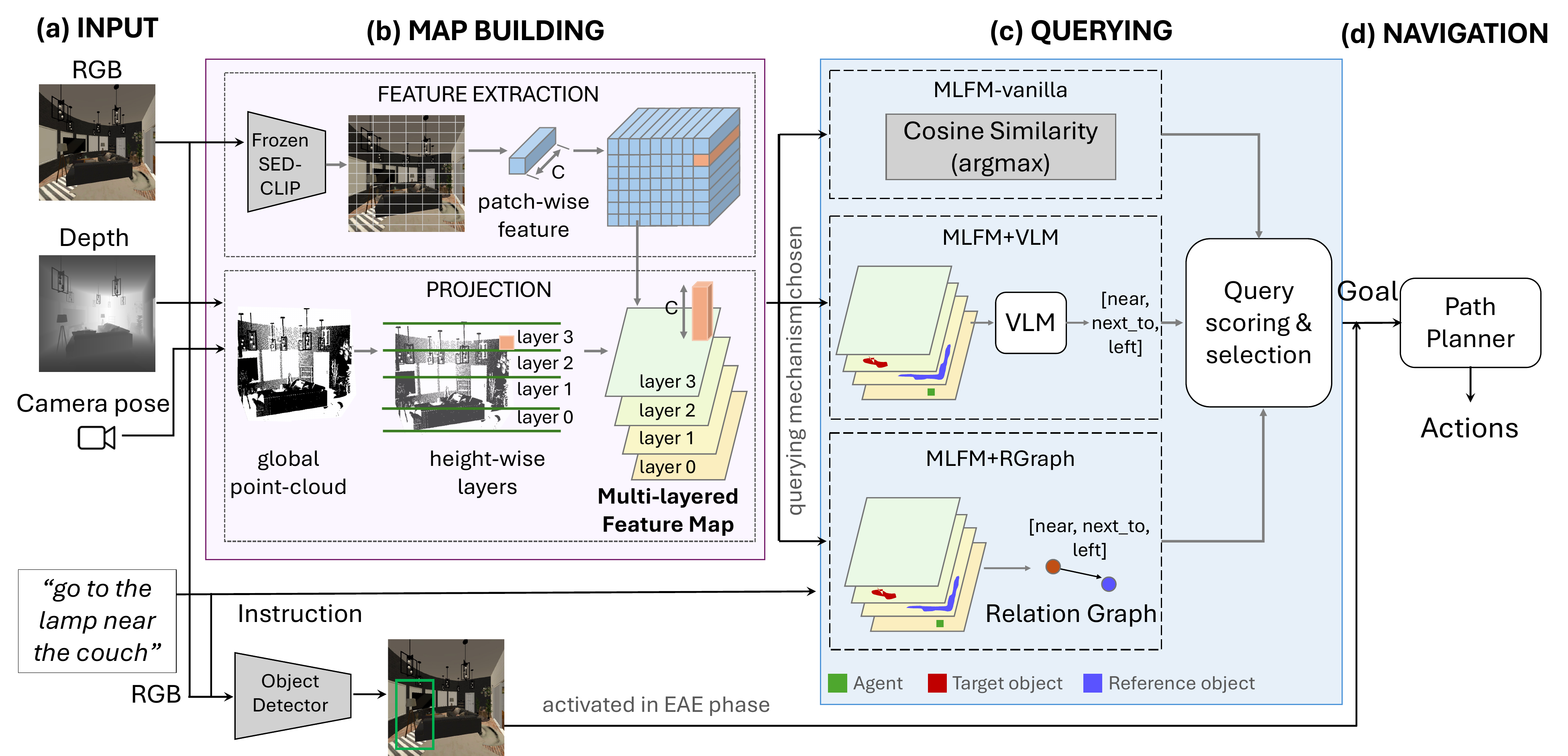}
  \caption{
\textbf{Method.} (a)~The agent takes as input the RGB image from which the map building~(b) extracts learned visual embeddings and projects onto layers using the depth and camera pose inputs. The map is then queried based on the input instruction
% We build a multi-layered map where each layer contains a 2D top-down projection with each cell storing learned visual embeddings. 
% We then query this map based on the input instruction 
by employing one of three techniques~(c)--\emph{vanilla}, \emph{VLM} or \emph{RGraph}. Once the agent identifies a possible goal location, it navigates to it using a path planner~(d). The agent activates the object detector as an additional signal during the initial phase (EAE) of the navigation.
  }
  \label{fig:map_building}
  \vspace{-15pt}
\end{figure}
To this end, we build a multi-layer feature map, $\mathcal{M}\in \mathbb{R}^{L  \times h  \times w \times f_d}$ where $L$ is the number of layers, each containing a 2D top-down map $m\in \mathbb{R}^{h  \times w \times f_d}$ and $f_d$ is the feature dimension stored at each map cell. $h$ and $w$ correspond to the ($X$ x $Y$) space of the physical environment, while $L$ corresponds to its discretized height dimension $Z$. 
\Cref{fig:map_building} presents a schematic of the map building process.
Given the RGB observation $I_t \in \mathbb{R}^{H  \times W  \times 3}$ at each step $t$, we extract patch-wise features $Fp_t \in \mathbb{R}^{H  \times W  \times n_p \times f_d}$ using CLIP~\citep{radford2021learning} image encoder from SED~\citep{xie2024sed}. We rely on a CLIP-based vision–language model to store features embedded in a joint image–text space, so that the map can be queried with natural language descriptions.
Here $n_p$ is the number of patches.
Using the depth observation $D_t \in \mathbb{R}^{H  \times W}$ and the camera intrinsics, we then project the features $Fp_t$ onto a 3D point-cloud.
Next we project the 3D points onto the 2D plane. For each point $p_i$ in the point-cloud at the 3D location ($x_i,y_i,z_i$), we find the appropriate layer using the height $z_i$ of the point in the point-cloud, such that: $l_i = \left\lfloor \frac{h_i}{\Delta h} \right\rfloor$, where $l_i$ is the layer index and $\Delta h$ denotes the height range of each layer.

\subsection{\newadd{Querying}}
We implement three different querying techniques for \methodnameshort to query the built map with the input goal description. Below we describe each technique.

\mypara{\mlfmvanilla.} This variant relies on the most basic querying technique to find the target on the multi-layer map by identifying candidates on the map where the cosine similarity between stored visual features and the query text is above a threshold and then selects the highest score as the target.

\mypara{\mlfmvlm.} In this variant, we use a large vision-language model (VLM) to query the relationship between objects. We first find the most likely locations of the target object and the reference object by computing the highest similarity scores for each. We then color the target object with `red' and the reference object with `blue'. We additionally mark the agent location on the map with a `green' square. We generate an image with image tiles corresponding to each layer of the map and indicate that tile 1 is at a lower height than tile 2, which is at a lower height than tile 3 and so on. We then prompt GPT-4 and ask it to infer the spatial relation between the `red' object A and the `blue' object B, from the agent's viewpoint:
\emph{``Infer the spatial relation between red A and blue B. If a green box is present, it indicates agent location. Decide relation from the agent's viewpoint.''}.
We compare the inferred relation to the query relation by encoding them with a CLIP text encoder and computing the cosine similarity. If the score is above a threshold, we consider the target as found.

\mypara{\mlfmrgraph.} 
In this variant we construct a \emph{relation graph} (\emph{RGraph}), with nodes representing objects and directed edges encoding spatial relations between object pairs (e.g., edge from A to B with label `above' indicates `A is above B'). Each node corresponds to a candidate location of either the target or the reference object, identified from the multi-layered map if its similarity score with the text query exceeds a threshold. Along with the object identity, we associate each node with the visual features extracted from its corresponding map region.
Edges are introduced between two nodes if the objects are in close spatial proximity, determined by the Euclidean distance on the map. Each edge is labeled with one or more relations inferred from the object map locations, and stores the corresponding CLIP text features. Note that this supports multiple relations between A and B, since A can be near B and at the same time to the left of B. 
In this way, the RGraph explicitly captures not only objects but also pairwise relations grounded in spatial structure and allows the agent to disambiguate which specific object pair is referenced (e.g. `the picture next to the cabinet' vs. `the picture above the cabinet'). See \Cref{sec:rgraph_more} for more details on the graph construction. 

\section{Experiments}
\label{sec:experiments}
In this section, we compare \methodnameshort with state-of-the-art zero-shot semantic navigation methods on our \DATASET dataset.
% we present a comprehensive evaluation of state-of-the-art zero-shot semantic navigation methods on our \DATASET dataset and compare with \methodnameshort.
% (see supplementary for GOAT-Bench results).

\mypara{Implementation.}
For all the baselines, we use an A*~\citep{hart1968formal, busch2024mapallrealtimeopenvocabulary} planner. 
% The agent take as inputs egocentric RGB-D images as well as GPS and compass readings relative to the starting pose in the episode. It also has access to the language instruction describing the current goal $g_i$ at any given time step. The action space is composed of four actions: \textit{move forward} by 25 cm, \textit{turn left} by 30°, \textit{turn right} by 30°, and \textit{found}. An episode is considered successful if the agent generates a \textit{found} within 1.5 meters of the goal.
% The agent is given a budget of 500 steps to navigate to each goal in an ordered sequence of three goals. The episode continues even when the agent fails to navigate to one goal.
We use SED~\citep{xie2024sed} patch features to build the multi-layer map and Yolo-World as an object detector in \methodnameshort. For experiments reported in this section, our map has three layers and each 2D grid cell side maps to 6 cm in the physical space. Using 1$\times$A40 GPU, it takes $\sim$8 hours for evaluating our method on the full test split.
For our experiments, we report the mean performance over five runs with random seeds and observe a standard deviation of $\pm$2.5.
% Note that we ran each of OneMap-v2, VLFM-v2 and \methodnameshort five times with random seeds and observe a standard deviation of $\pm$2.5 for each. In the tables, we report the mean performance over five runs.

\mypara{Evaluation metrics.}
% We evaluate how the semantic navigation methods perform across the various aspects of the language descriptions. 
We use standard navigation metrics, the success rate (\textit{SR}) and the success weighted by inverse Path Length (\textit{SPL})~\citep{anderson2018evaluation} to measure whether the agent can reach a target object successfully and how well the agent trajectory matches the shortest path to the target respectively. More specifically, we report SR against each of the linguistic features--attributes (\emph{color, size}, etc.) and spatial relations (\emph{support, egocentric directional}, etc.). 
% We follow the task setup from GOAT-Bench, where the agent is given a time budget of 500 steps to navigate to each goal in a sequence of multiple goals, and the episode continues even when the agent fails to navigate to one goal.

% \subsection{Baselines}
% We compare to two SOTA methods from the ObjectNav task (VLMaps \& VLFM), that uses 2D semantic mapping, as our baselines. We adapt each method to the MultiON task by iteratively applying it to each goal object and retaining the built feature map across goals in an episode. From the MultiON SOTA methods, we use OneMap.
\mypara{Baselines.}
We focus on understanding how an explicit semantic map helps agents solve the \taskshort task on the \DATASET dataset, especially when the goal descriptions vary in linguistic attributes.
Accordingly, we compare \methodnameshort\ with state-of-the-art \emph{mapping-based} zero-shot navigation methods: \textbf{VLMaps}~\citep{huang2023visual}, \textbf{VLFM}~\citep{yokoyama2023vlfm}, \textbf{OneMap}~\citep{busch2024mapallrealtimeopenvocabulary} and \textbf{MOPA}~\citep{Raychaudhuri_2024_WACV} (see \Cref{sec:baselines_more}).
% For VLMaps and VLFM we extend the original ObjectNav code to the multi-goal setting by running their policy sequentially for each target while preserving the feature map across goals, so that every method benefits from the same accumulated memory within an episode.
We have two additional baselines, \textbf{VLFM-v2} and \textbf{OneMap-v2}, where we adapt the two phase-navigation EAE-E (\Cref{sec:phases}), explore-and-exploit (EAE) followed by exploit-only (E) phases, for the two baselines, VLFM and OneMap. 
\newadd{
Furthermore, we compare with a baseline \textbf{\egoimagemap}, where we store egocentric RGB images on the map, similar to Modular GOAT~\citep{chang2023goat,khanna2024goat}. We then use the two phase-navigation (EAE-E) and query GPT-5 with the stored images to select the goal.
}

\subsection{Results on \DATASET}
Below we discuss agent performance on various aspects of language understanding. \Cref{tab:results_main} shows that \methodnameshort variants (rows 8-10) outperform the 2D map based baselines (rows 1-7) on overall SR and SPL, highlighting the effectiveness of the granular multi-layered representation over 2D top-down maps. This applies to fine-grained attribute understanding (with few exceptions), spatial understanding, as well as descriptions containing only goal category (+3.9\% gain in row 8 over 6). 
% Below we report our findings in detail.

% \subsubsection{\newadd{Attribute understanding}}

\mypara{Multi-layered maps capture fine-grained linguistic cues better than 2D maps.}
\Cref{tab:results_main} shows that \methodnameshort significantly outperforms the baselines (rows 8-10 vs. 1-7) in attribute understanding, especially on attributes--color, number, material and modifier--by +3.8\%, +7.2\%, +51.6\% and +2.8\% respectively (row 8 vs. 6). This indicates that the layered map represents visual features better than a 2D map, where the projected features get aggregated on a single layer, making it hard to distinguish intricate details about objects. Moreover, \mlfmrgraph outperforms \mlfmvanilla and \mlfmvlm on color, material and modifier attributes.
However, \methodnameshort fails to identify texture of objects. This is mainly the limitation of the feature extractor (see ablation in \Cref{sec:ablation_feature}). \methodnameshort also fell short of OneMap-v2 on descriptions with the state attribute (see analysis is \Cref{sec:failure_mlfm_onemap}).
% Note that SR for attribute understanding in rows 9-10 are the same as the \mlfmvanilla, since those two variants only differ from the vanilla in querying with respect to spatial relations. For attribute understanding, they rely on vanilla querying method, which returns the highest scoring location in the map. For example, in \mlfmvlm we give images to the VLM with abstract blobs representing high similarity areas on the top-down map, asking it to infer the relations between the two. In case the image has one blob, which has no attribute-level encoding, the VLM returns ``undetermined''.
% \methodnameshort achieves +30.3\% increase in SR and +11.5\% increase in SPL over OneMap and +40.6\% increase in SR and +15.9\% increase in SPL over VLFM.

\mypara{Two-phase navigation is effective in long-horizon tasks.}
Poor performance in VLFM can be attributed to the reliance on the built map, even during the initial stages of navigation when the map may not contain the target object. This is evident from the improved performance of VLFM-v2 (row 5),
% (+21.8\% and +8.7\% increase in SR and SPL respectively) 
where we adapt the two-phase navigation, over the vanilla VLFM (row 2), even when both use 2D maps. We observe a similar trend in OneMap (rows 6 vs. 4).
% , where OneMap-v2 achieves +26.1\% and +9.7\% increase in SR and SPL respectively. 
This shows that our two-phase navigation (EAE-E) approach is extremely effective when dealing with longer-horizon tasks.

\begin{table}[t]
\centering
\vspace{-10pt}
\caption{
     \textbf{\newadd{Performance.}} 
     \mlfmrgraph outperforms others on overall metrics, most attributes and spatial relations.
     The \methodnameshort variants do best on `cat' and on `color', `num', `mat' and `mod' attributes. 
     \small{Note-\emph{cat}:category, \emph{tex}:texture, \emph{num}:number, \emph{mat}:material, \emph{mod}:modifier, \emph{ego-dir}:egocentric directional, \emph{allo-dir}:allocentric directional, \emph{supp}:support, \emph{prox}:proximity, \emph{cont}:containment.}
     }
\label{tab:results_main}
\resizebox{\linewidth}{!}{
\begin{tabular}{lrrrrrrrrrrrrrrrrr}
\toprule
Methods &\multicolumn{2}{c}{Overall} &  & &\multicolumn{7}{c}{Attribute understanding} & &\multicolumn{5}{c}{Spatial relations understanding} \\
\cmidrule{2-3} \cmidrule{6-12} \cmidrule{14-18}
&SR$\uparrow$ &SPL$\uparrow$ &cat & &color &size &tex &num  &mat &state &mod & 
&\thead{ego-\\dir} &\thead{allo-\\dir} &supp  &prox &cont
\\
\midrule

1) VLMaps &5.0 &1.2 &3.1 &&1.4 &0.0 &0.0 &0.0 &0.0 &0.0 &3.0 & &0.0 & 1.0 &0.6 &3.7 &0.0 \\
2) VLFM &7.1 &2.6 &2.4 &&1.6 &0.0 &0.0 &0.0 &0.0 &0.0 &3.3 &  &0.4 &1.6 &1.3 &6.9 &2.1  \\
3) MOPA &6.3 &4.4 &2.8 &&2.6 &3.6 &0.0  &0.0  &0.0  &0.0 & 4.2 &&0.0 & 2.1 &1.4 &4.1 & 2.0 \\ 

4) OneMap &15.3 & 7.1 &3.7 &&4.7 &3.9 &3.0 &2.3 &3.1 &0.0 & 6.7 & &2.1 & 4.1 &10.7 & 5.5 &5.7 \\

\midrule
5) VLFM-v2 &24.8 & 9.7 &28.2 && 11.5 & 33.3 &0.0  &7.1 &43.1  &28.6 &17.7 & &0.0 &2.8 &12.9 &4.2 &3.3   \\

6) OneMap-v2 &26.7 &10.1 &29.1 &&26.9 &\textbf{41.7} &0.0 &7.1 &0.0  &\textbf{57.1} &21.1 & &2.1 &4.7 &22.1 &5.9 &6.5    \\

7) \egoimagemap &35.1 &10.9 &\textbf{33.7} && 34.8 &39.5 &\textbf{33.3} &11.7 &53.1 &17.8 &35.0 &&34.6 &20.2 &33.3 &22.2 & 17.5 \\

\midrule

8) \mlfmvanilla &28.8 &10.3
&33.0 &&30.7 &\textbf{41.7} &0.0 &\textbf{14.3}  &51.6 &14.3 &23.9 &&16.7 &12.5 &20.4 &14.1  &14.3  \\

9) \mlfmvlm  &33.3 &10.7 
&33.3 &&34.5 &\textbf{41.7} &0.0 &\textbf{14.3} &51.9 &14.3 &31.3
% &\textbf{33.0} & &\textbf{30.7} &\textbf{41.7} &0.0 &\textbf{14.3}  &\textbf{51.6} &14.3 &\textbf{23.9} 
&&17.1 & 19.3 &21.2 & 16.9 & 14.7 \\

10) \mlfmrgraph  &\textbf{39.5} &\textbf{14.9} 
&33.6 &&\textbf{34.9} &\textbf{41.7} &0.0 &\textbf{14.3} &\textbf{53.8} &14.3 &\textbf{37.6} 
% &\textbf{33.0} &&\textbf{30.7} &\textbf{41.7} &0.0 &\textbf{14.3}  &\textbf{51.6} &14.3 &\textbf{23.9} 
&& \textbf{37.7} &\textbf{20.7} &\textbf{34.2} & \textbf{32.5} & \textbf{21.1}  \\

\bottomrule

\end{tabular}
}
\vspace{-12pt}
\end{table}

% \subsubsection{\newadd{Spatial relation understanding}}
\mypara{Multi-layered maps capture spatial relations better than 2D maps.}
% \newadd{
\Cref{tab:results_main} shows that \emph{\mlfmvanilla} with its basic querying mechanism outperforms VLFM-v2 and OneMap-v2 (rows 8 vs. 5-6) across all spatial relation types. This reaffirms that the multi-layer map not only helps in attribute understanding, but the granular representation also helps in spatial understanding.

\mypara{Spatial understanding benefits from enhanced representation and querying.}
Both \emph{\mlfmvlm} (row 9) and \emph{\mlfmrgraph} (row 10), equipped with enhanced querying mechanisms, outperform \emph{\mlfmvanilla} across all relation types, with \emph{\mlfmrgraph} achieving the best performance. This underscores that spatial relationships can be effectively modeled as graphs, allowing for better reasoning capabilities.

\mypara{VLMs reason better on raw egocentric images than top-down abstract projections.}
\emph{\egoimagemap} (row 7) achieves comparable performance as our best performing model, and outperforms \emph{\mlfmvlm}. This indicates that the VLM is able to reason over egocentric RGB images better than top-down projections of image features (see \Cref{sec:failure_mlfm_onemap}). Moreover, it achieves the best results on descriptions with only goal category and with texture attribute.
% Notably, \emph{\mlfmrgraph} achieves a +3.3\% gain in overall SR over \emph{\mlfmvlm}. Nevertheless, the overall success rate remains at 20\%, highlighting substantial room for improvement.
% \input{tables/iclr/tab-results-spatial-main}

\subsection{Failure analysis}
\mypara{\methodnameshort vs. OneMap-v2.}\label{sec:failure_mlfm_onemap} To understand why OneMap-v2 performs better than \mlfmvanilla for descriptions with \emph{state} attribute, 
% we look at the error cases in \Cref{tab:failure_mlfm_vs_onemap}. 
we compute the percentage of times the agent makes a wrong detection or identifies the wrong goal on the map or runs out of time. 
First, we observe that there are more wrong detections (71.4\% in \mlfmvanilla vs. 28.6\% in OneMap-v2) than wrong goals on the map (14.3\% in both). This indicates that performance could be improved by using a better object detector (confirmed by ablation in \Cref{sec:ablation_detector}).
Second, \mlfmvanilla and OneMap-v2 have the same wrong goal on map percentage (14.3\%), indicating that this error is not due to using multi-layer map instead of 2D map. Moreover, an ablation in \Cref{sec:ablation_feature} indicates that the performance in \emph{state} attribute improves by using image-level features instead of patch-level features.

\mypara{Overall \methodnameshort failure.}
Performing a similar failure analysis on the overall performance of \mlfmvanilla, we observe that 31.7\% of the error cases are due to wrong detection and 17.0\% are due to wrong goal identification on the map. The remaining 48.3\% is due to the agent running out of time budget while exploring and 3\% is where the agent identifies the goal but fails to reach it on time.

\begin{figure}[t]
\centering
\includegraphics[trim={0 1.5cm 0 1cm},width=\linewidth]{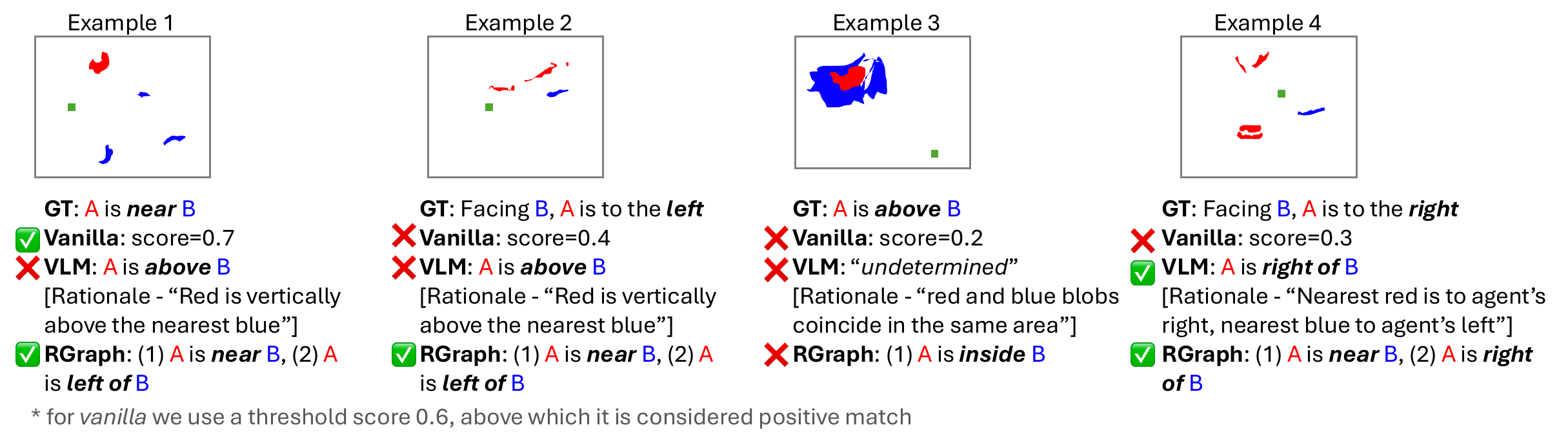}
\vspace{-8pt}
\caption{
Comparisons showing that \emph{VLM} struggles to reason on projected abstract features, often interpreting them as egocentric views (example 1). \emph{RGraph} struggles distinguishing `inside' from `above' when both objects are projected onto the same map layer (example 3). 
}
\label{fig:spatial_result_compare}
\vspace{-15pt}
\end{figure}

\mypara{\newadd{\mlfmvlm vs. \mlfmrgraph.}}
We find that the primary source of failure for \emph{\mlfmvlm} arises from the bias of VLMs, which are trained on web-scale photo and video corpora dominated by egocentric perspectives. As a result, they infer spatial relations more reliably from egocentric images than from top-down maps. For instance, when presented with a top-down projection where A's red blob appears above B's blue blob, the VLM tends to predict `A is above B', mirroring how it would interpret an egocentric image, rather than the intended relation A is `near' or `next to' B (see \Cref{fig:spatial_result_compare}).
\emph{\mlfmrgraph}, on the other hand, might fail to differentiate between `inside', `above', `below' and `on top of' if both the objects are projected on the same map layer, depending on the number of layers in the multi-layered map.

% \section{Ablations}
% \label{sec:ablations}

\subsection{Ablations}
\label{sec:ablation_feature}

To gain more insights into why \methodnameshort fails to understand attributes like \emph{texture} and \emph{state}, we perform ablations (\Cref{tab:results_features_detector}) with open-vocabulary feature extractors and object detectors.

\mypara{Patch vs. image vs. pixel features.} When compared with their image-level counterparts (rows 1–2 vs.\ 3–4), SED-CLIP and BLIP-2 patches generally perform better on attribute understanding (with a few exceptions).
% raise SR by \(\,+11.8\%\) and \(\,+8.0\%\), and SPL by \(\,+4.7\%\) and \(\,+0.5\%\), respectively.
% The explanation is twofold.  
% First, 
This can be attributed to the fact that when we store the same global feature vector in every layer (CLIP or BLIP image-level), the map gains no additional information.  
% Second, SED~\citep{xie2024sed} and BLIP-2 patch embeddings preserve spatial alignment as well as finer image–text correspondence.
This is evident for 
% support relations (\(+6.4\%\)) and for 
\emph{color}, \emph{size}, \emph{material}, and \emph{modifier} cues (\(+26.9\%, +25.0\%, +16.3\%, +3.1\%\), respective gains in row 2 over 4). 
% and even for descriptions without explicit attributes (\(+1.0\%\)).  
% \mypara{Image features are sometimes useful.}
Image features, on the other hand, edge out patches on \textit{number} and \textit{state} attributes (rows 2 vs.\ 4), suggesting that a holistic view occasionally helps when counting instances or detecting object states.  
% BLIP-2 outperforms CLIP on texture and number recognition (row 4 vs.\ 3), confirming its stronger vision–language alignment for such properties.
% \mypara{Pixel features fail to capture object-level understanding.}
Pixel features from LSeg perform worst overall (row 5), indicating that overly fine-grained features hinder object-level reasoning and dilute broader context. 
% captured by patch- or image-level representations.

\mypara{BLIP-2 vs. CLIP.}
BLIP-2 patches outperforms SED-CLIP patches on the \emph{color} (+3.9\%) and \emph{modifier} (+1.0\%) attributes (row 2 vs.\ 1). Additionally, BLIP-2 image outperforms CLIP on \emph{texture} (+20.0\%) and \emph{number} (+21.4\%) comprehension (row 4 vs.\ 3). These findings confirm BLIP-2's stronger vision-language alignment for such properties.

\begin{table}[ht]
\vspace{-8pt}
\caption{
     \textbf{Ablations.}
     Patch outperforms image followed by pixel features. 
     % Image-level features, however, perform better when there are no attributes in the descriptions or the ones with state attribute. 
     However, the choice of feature extractor and object detector might influence understanding of certain attribute types.
     % Grounding-DINO outperforms Yolo-World in size and state attributes and descriptions with no attributes.
     }
\label{tab:results_features_detector}
\centering
\resizebox{\linewidth}{!}{
\begin{tabular}{lllcrrrrrrr}
\toprule
\multirow{2}{*}{Object detector} &\multirow{2}{*}{Features} &\multirow{2}{*}{Type} & &\multicolumn{7}{c}{Attribute understanding} \\
\cmidrule{5-11}
&&& &color &size &texture &number &material &state &modifier  \\
\toprule

1) YOLO-World~\citep{cheng2024yolo} & SED~\citep{xie2024sed} & patch & &30.7 &41.7 &0.0 &14.3 &\textbf{51.6} &14.3 &23.9 \\

2) & BLIP-2~\citep{li2023blip} & patch & &\textbf{34.6} &41.7 & \textbf{20.0} & 14.3 &16.3 &14.3 &24.9  \\

\cmidrule{2-11}

3) &CLIP~\citep{radford2021learning} & image & &11.5 &8.3 & 0.0 &0.0 &0.0 &28.6 &17.1  \\

4) &BLIP-2~\citep{li2023blip} & image & &7.7 &16.7 &\textbf{20.0} &\textbf{21.4} &0.0 &28.6 &21.8 \\

\cmidrule{2-11}

5) &LSeg~\citep{li2022language} & pixel &  &3.8 &0.0 &0.0 &7.1 &0.0 &14.3 &7.1 \\

\midrule

6) Grounding-DINO~\citep{liu2024grounding}  & SED~\citep{xie2024sed} & patch & &26.9 &\textbf{49.7} &0.0 &0.0 & 53.7 &\textbf{42.8} &\textbf{25.7} \\

\bottomrule
\end{tabular}}
\vspace{-5pt}
\end{table}

% \subsection{Ablation on object detectors}
\mypara{Grounding-DINO vs. YOLO-World.}
\label{sec:ablation_detector}
To assess the influence of the open-set detector, we swap YOLO-World for Grounding-DINO inside \methodnameshort\ (rows 1 vs. 6).  
% YOLO-World achieves slightly higher overall scores (\,+0.9 \% SR, +1.7\% SPL), yet
Grounding-DINO ends up stronger for specific attribute types, such as +8.0\% for \textit{size}, +28.5\% for \textit{state} and +1.8\% for \emph{modifier}.

Overall, we find that patch features are often better at capturing granular attribute cues than pixel or image features. However the choice of specific feature extractors and object detectors might heavily influence understanding of certain attribute types.

\subsection{Results on GOAT-Bench}
\label{sec:goat_performance}
To demonstrate the effectiveness of our \emph{\mlfmrgraph} method on real-world scans, we perform an experiment on the language goals from GOAT-Bench~\citep{khanna2024goat} and compare to three baselines from the GOAT-Bench paper -- the zero-shot \emph{Modular GOAT}~\citep{chang2023goat} and two trained methods: \emph{RL Skill Chain} and \emph{RL Monolithic}. 
% \mypara{Task.} 
% Similar to our task, GOAT-Bench requires the agent to navigate to multiple goals, although the number of goals in a single episode range from 5 to 10, compared to 3 in our task. Moreover, although the goals are described using object category, language description or image of the object instance,
% In the GOAT-Bench task, an agent is required to sequentially navigate to multiple goals, described by the object category, a language description or an image of the object instance. 
% in this paper, we focus only on the language goals where the goal object is described using a detailed description (e.g. `a black leather couch next to coffee table').
% \mypara{Baselines.}
% We compare \mlfmrgraph to three baselines from the GOAT-Bench paper -- the zero-shot \emph{Modular Goat}~\citep{chang2023goat} and two trained methods \emph{RL Skill Chain} and \emph{RL Monolithic} are trained methods. 
% In RL Skill Chain, individual navigation policies were trained on three different skills, ObjectNav, InstanceImageNav and LanguageNav, to tackle object category, image goal and language goals respectively. These policies are then combined using a third policy for high-level planner. On the other hand, RL Monolithic was trained end-to-end by using a multimodal goal encoder that encodes goals from different modalities into a common latent space. 
\mlfmrgraph, being a zero-shot method is however only comparable to the Modular GOAT method, that builds a semantic map memory (containing object categories) as well as an instance-specific map memory (containing egocentric images and CLIP image features).
% \mypara{Results.}
\Cref{tab:results_goat_bench} shows that \mlfmrgraph outperforms the baselines 
% including the zero-shot method and the trained methods in both SR and SPL, 
on the language goals of GOAT-Bench, with +2.7\% in SR and +5.3\% in SPL.

\begin{table}[t]
% \vspace{-10pt}
\caption{
 On GOAT-Bench, \mlfmrgraph outperforms the baselines on language goals.
}
\label{tab:results_goat_bench}
\centering
    \begin{tabular}{lcccc}
    \toprule
    Methods &trained &zero-shot &SR &SPL  \\
    \midrule

    \rowcolor{gainsboro}
    RL Skill Chain &$\checkmark$ &$\times$ &16.3 &7.4 \\

    \rowcolor{gainsboro}
    RL Monolithic &$\checkmark$ &$\times$ &12.6 &6.5 \\
    
    Modular GOAT &$\times$ &$\checkmark$ &21.5 &16.2 \\

    \midrule
    
    \mlfmrgraph &$\times$ &$\checkmark$ &\textbf{24.2} &\textbf{21.5} \\

    \bottomrule
    \end{tabular}
    \vspace{-10pt}
\end{table}

\section{Conclusion}
\label{sec:conclusion}
We extend the semantic navigation task to \emph{\taskshort}, to evaluate attribute-aware and spatially-aware language understanding.  
We propose \emph{\DATASET} dataset to implement this task, where episodes have sequences of goals, with manually verified descriptions and per-instruction linguistic tags.
We also propose a novel method, \emph{\methodname} (\emph{\methodnameshort}), that preserves fine-grained visual detail in a multi-layer semantic map representation and we demonstrate that it improves fine-grained and spatial language grounding. 
% Systematic experiments and ablations show that \methodnameshort\ surpasses state-of-the-art mapping baselines and that patch-level CLIP or BLIP-2 features, when stored across height slices, yield the best performance.
Experiments show that language understanding is still far from being solved and offer ample scope for new ideas to improve perception, memory, and language grounding.
% We also discussed the limitations of our dataset in terms of linguistic patterns and spatial relationships, which could enable future work in this direction. 
We hope our findings provide useful insights to drive future research in this direction.
% into natural language understanding in embodied AI agents.

\mypara{Limitations.}
\label{sec:limitations}
The \DATASET\ dataset has several limitations including lacking complex linguistic structures such as coreference (``place \emph{it} on the table''), negation (``a chair that is \emph{not} black''), interaction-based action directives (``pick up'', ``bring back'') and multi-object relations (``chairs around the dining table'') (see \Cref{sec:limitations_more}). However, we leave this for future work as possible extensions to \DATASET.

\mypara{LLM usage.} An LLM was used in the dataset generation process (see \Cref{sec:dataset} and \Cref{sec:supp_llm_usage} for detail), in the \egoimagemap baseline (see \Cref{sec:method} and \Cref{supp:llm-egoimage}) and in the \mlfmvlm method (see  \Cref{sec:method} and \Cref{supp:llm-mlfmvlm}). We also sparingly used an LLM~\citep{openai2023gpt4} for rewording some text in the paper.

\paragraph{Acknowledgement.}

This work was funded in part by a CIFAR AI Chair, a Canada Research Chair, and NSERC Discovery grants. Infrastructure was supported by CFI JELF grants. TC was supported by the PNRR project Future AI Research (FAIR - PE00000013), under the NRRP MUR program funded by the NextGenerationEU.
We thank Jelin Raphael Akkara for sharing his viewpoint generation code to be used in this project.
We also thank Francesco Taioli and Austin T. Wang for helpful discussions and feedback.

\mypara{Contributions.}

\mypara{\bf Sonia Raychaudhuri} created the \DATASET dataset, implemented the \methodnameshort codebase, re-implemented and integrated baseline methods, conducted all experiments reported in the paper, and contributed extensively to writing the paper.

\mypara{\bf Enrico Cancelli} developed the initial codebase and performed preliminary experiments.

\mypara{\bf Tommaso Campari} contributed to the refinement of ideas, coordinated the team, and was extensively involved in paper writing.

\mypara{\bf Lamberto Ballan, Manolis Savva and Angel X. Chang} provided supervision, guidance throughout the project, and critical feedback on the manuscript.

{
    \small
    \bibliographystyle{plainnat}
    \bibliography{bibliography}
}

\clearpage
\appendix

\section{Supplementary Material}
\label{sec:appendix}
% In this supplementary document, we provide more details on the \DATASET dataset (\Cref{sec:dataset_more}), the viewpoint generation approach that we follow while storing viewpoints in the episodes (\Cref{sec:viewpoint_gen}), more ablations to determine how the \methodnameshort components contribute to the agent performance (\Cref{sec:ablations_more}) and finally the performance of \methodnameshort on the GOAT-Bench language goals (\Cref{sec:goat_performance}).
In this supplementary document we
(i) present additional details of the \DATASET\ corpus (Sec.~\ref{sec:dataset_more});
(ii) provide more detail on the method (Sec.~\ref{supp:method-more});
% describe the procedure used to generate and store episode viewpoints (Sec.~\ref{sec:viewpoint_gen});
(iii) briefly describe the baselines used in the main paper (Sec.~\ref{sec:baselines_more});
and 
(iv) report further ablations that quantify the contribution of each component in our method (Sec.~\ref{sec:ablations_more}).
% and
% (iv) evaluate \methodnameshort\ on the GOAT-Bench language-goal split (Sec.~\ref{sec:goat_performance}).

\subsection{\DATASET dataset}
\label{sec:dataset_more}
In this section we present more details about our \DATASET dataset, including details on dataset statistics (\Cref{tab:dataset_stats}).

\subsubsection{LLM for dataset generation}
\label{sec:supp_llm_usage}

\mypara{Forming coherent sentences from annotations.}
We prompt GPT-4~\citep{openai2023gpt4} to form a sentence in natural language that would make sense to a human:

\begin{quote}
    ``\texttt{You will be provided with a list of attributes, and your task is to convert them to a free-flowing natural language sentence in English as spoken by native speakers. Drop brand names and all capital lettered words from the text.}''
\end{quote}

We also provided a list of five examples to improve GPT's performance via in-context learning.

\mypara{Extracting attributes from sentences for annotations.}
We use GPT-4 to also extract attributes from a language sentence which we then store as the fine-grained linguistic annotations, so that we can evaluate agent's performance on language understanding:

\begin{quote}
    \texttt{You are given a sentence with the description of an object that someone is supposed to find in a scene. Your goal is to extract the attributes used to describe the object: <input\_sentence>. Return the output in a JSON format according to the following format:}
\end{quote}

\colorlet{punct}{red!60!black}
\definecolor{background}{HTML}{EEEEEE}
\definecolor{delim}{RGB}{20,105,176}
% \colorlet{numb}{magenta!60!black}

\lstdefinelanguage{json}{
    basicstyle=\small\ttfamily,
    showstringspaces=false,
    breaklines=true,
    literate=
      {,}{{{\color{punct}{,}}}}{1}
      {\{}{{{\color{delim}{\{}}}}{1}
      {\}}{{{\color{delim}{\}}}}}{1}
      {[}{{{\color{delim}{[}}}}{1}
      {]}{{{\color{delim}{]}}}}{1},
}

\begin{lstlisting}[language=json]
        {"attribute_types": {
        "color": {
            "exists": True if attribute is found in prompt or False otherwise,
            "explanation": list of attributes identified, or empty if none
        },
        "size": {
            "exists": True if attribute is found in prompt or False otherwise,
            "explanation": list of attributes identified, or empty if none
        },
        "shape": {
            "exists": True if attribute is found in prompt or False otherwise,
            "explanation": list of attributes identified, or empty if none
        },
        "number": {
            "exists": True if attribute is found in prompt or False otherwise,
            "explanation": list of attributes identified, or empty if none
        },
        "material": {
            "exists": True if attribute is found in prompt or False otherwise,
            "explanation": list of attributes identified, or empty if none
        },
        "texture": {
            "exists": True if attribute is found in prompt or False otherwise,
            "explanation": list of attributes identified, or empty if none
        },
        "function": {
            "exists": True if attribute is found in prompt or False otherwise,
            "explanation": list of attributes identified, or empty if none
        },
        "style": {
            "exists": True if attribute is found in prompt or False otherwise,
            "explanation": list of attributes identified, or empty if none
        },
        "text_label": {
            "exists": True if attribute is found in prompt or False otherwise,
            "explanation": list of attributes identified, or empty if none
        },
        "state": {
            "exists": True if attribute is found in prompt or False otherwise,
            "explanation": list of attributes identified, or empty if none
        },
        "modifier": {
            "exists": True if attribute is found in prompt or False otherwise,
            "explanation": list of attributes identified, or empty if none
        }}}
\end{lstlisting}

\subsubsection{Episode generation}
For each episode in the \DATASET dataset, we first draw a random navigable start pose for the agent, then choose an object category that satisfies the following constraints:
\begin{inparaenum}[(i)]
\item at least one navigable viewpoint lies within 1.5 m of the object;
\item the start pose and all three goals are on the same floor;
\item every goal is reachable from the start pose;
\item each goal is at least 2 m from the start pose;
\item the geodesic-to-Euclidean distance ratio from the start pose to each goal exceeds~1, ensuring non-trivial navigation;
\item the geodesic distance from the start pose to the first goal is at least 2.5 m.
\end{inparaenum}
For each goal, we also store `viewpoints' or navigable positions around the goal object, such that navigating near to any of those will be considered successful (see \Cref{sec:viewpoint_gen}).

\begin{table}[t]
% \vspace{-8pt}
\caption{\textbf{\newadd{Dataset statistics.}}  
\DATASET contains val and test splits, each with distinct sets of scenes and goal object categories. The goal descriptions might contain one or more linguistic features.}
\label{tab:dataset_stats}
\centering
\begin{tabular}{lrrr}
\toprule
& Validation & Test & Overall \\
\toprule
scenes     &20    &15  &35  \\
episodes     &932 &875  &  1807 \\

distinct object categories &19 &12 &31 \\
distinct object instances &160 &154 &314 \\

goal descriptions     &2796  &2625 &5421 \\

descriptions with attributes/relationships &1659 &2181 &3840 \\

total linguistic features &2534 &3524 &6058 \\
unique linguistic feature values &136 &99 &235 \\
\bottomrule
\end{tabular}
% \vspace{-10pt}
\end{table}

\subsubsection{\newadd{Spatial relation descriptions}}
Here we describe in detail the procedure used to extract spatial relationships between a pair of objects from the ground-truth 3D object bounding boxes in HSSD.
\newadd{For \emph{egocentric directional relations}, we take two objects with non-overlapping bounding boxes and compute the centroids of each bounding box. We transform them into the camera coordinate frame using the provided camera extrinsics. We then project these centroids into the image plane using the camera intrinsics and compare their horizontal pixel coordinates to assign ``left'' or ``right'' relations. This ensures that directional relations are defined relative to the agent's pose and viewpoint.}
\newadd{For \emph{allocentric relations}, we directly compare the relative positions of object bounding boxes in the world coordinate system. For allocentric \emph{directional relations}, we cast rays from the object's bounding box in the direction of the relation (vertical direction for ``above'' and ``below'') to retrieve the object with hits\footnote{Code adapted from Habitat-Lab \url{https://github.com/facebookresearch/habitat-lab/blob/5e0d63838cf3f6c7008c9eed00610d556c46c1e3/habitat-lab/habitat/sims/habitat_simulator/sim_utilities.py\#L724}}. For instance, object $A$ is ``above'' $B$, if rays from $A$'s bounding box when shot downwards hit $B$'s bounding box. 
For allocentric \emph{support relations}, we utilize the ground-truth support data available for the objects in HSSD. 
Allocentric \emph{proximity relations} are derived by measuring the minimum Euclidean distance between the surfaces of two bounding boxes. We adopt a tiered thresholding scheme - objects within 0.0–0.2 meter are labeled as ``next to'' each other, while objects within 0.2–1.0 meter are labeled as ``near'' each other. This design captures both immediate adjacency and a more relaxed proximity. Finally, \emph{containment relations} are identified when the bounding box of one object is fully enclosed within that of another across all three spatial dimensions.}
\newadd{Our procedure benefits from the availability of ground-truth 3D bounding boxes in synthetic scenes, allowing us to generate accurate and consistent spatial relation annotations. The use of ray-casting provides a principled way to capture directional relations and the surface-to-surface distances enable a faithful encoding of adjacency to capture proximity relations. The pipeline is fully automatic and scalable, making it possible to construct a large dataset of spatial relations without requiring manual annotation.
}

\subsubsection{Understanding the dataset.}

\mypara{Why the need for a new dataset?}
% There are a few datasets~(\Cref{tab:dataset_comparison}) available in the community to evaluate ObjectNav~\citep{anderson2018evaluation,batra2020objectnav}. 
The closest to our dataset is the GOAT-Bench~\citep{khanna2024goat} which introduced a multi-object navigation dataset that combines ObjectNav and InstanceNav goal descriptions. In contrast to GOAT-Bench, our dataset contains language descriptions at varying amounts of specificity, such that one or multiple target locations could be a match. We also include annotations for linguistic tags for each description which allows for fine-grained language evaluation. Moreover, GOAT-Bench contains several errors propagated from the VLM, which we mitigate by using ground-truth object annotations and also manually validating them. 

\begin{figure}[t]
  \centering
  \includegraphics[width=\linewidth]{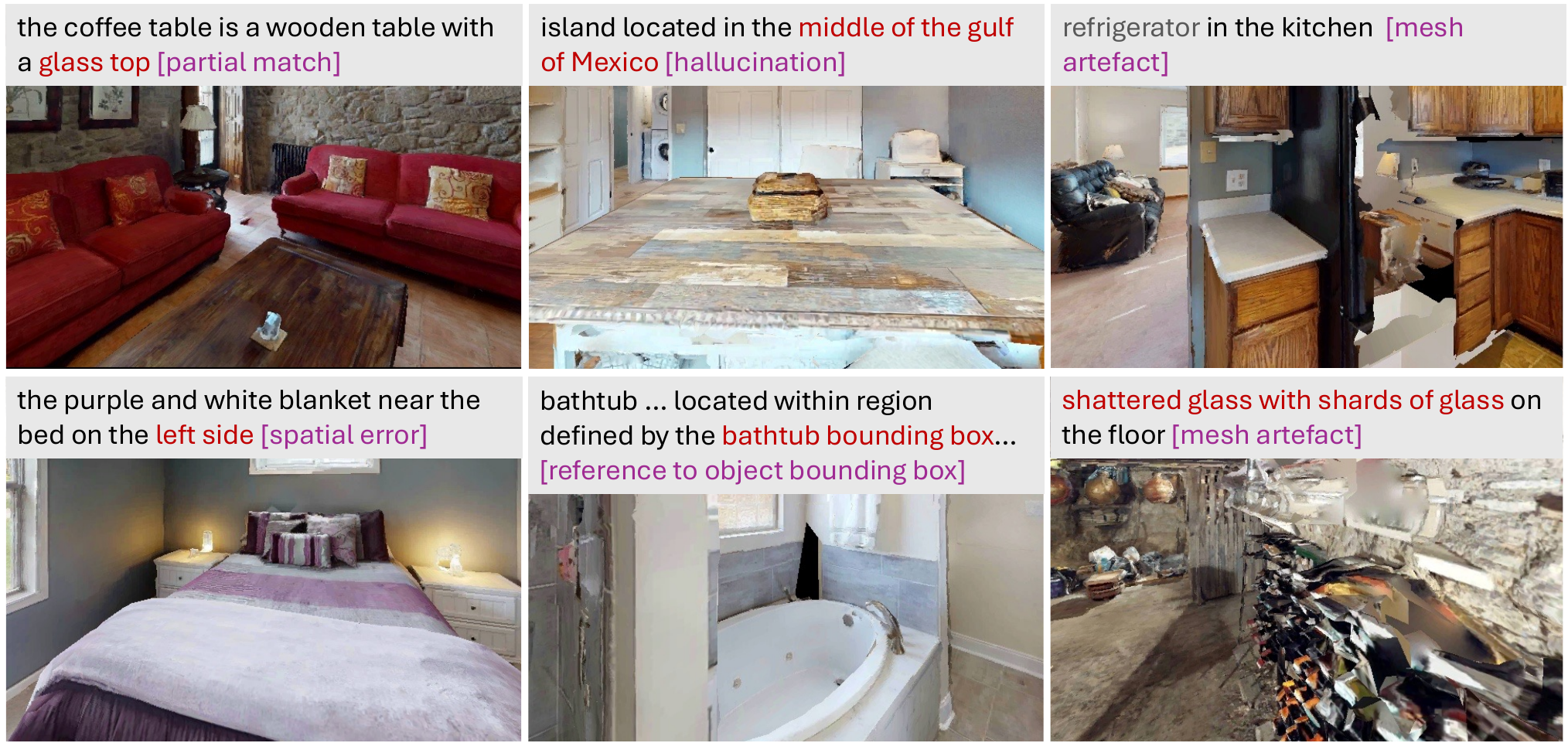}
  \caption{Language descriptions in Goat-Bench contain errors propagated from the BLIP-2 model. This figure shows examples for \emph{partial match}, \emph{hallucination}, \emph{mesh artifact}, \emph{spatial error} and \emph{reference to object bounding box} errors.
  }
  \label{fig:goat_errors}
\end{figure}

The dataset creation pipeline in GOAT-Bench derives object attributes by prompting a pretrained BLIP-2 model, and this reliance on automatic extraction introduces noticeable noise in the captions. Typical issues include (\Cref{fig:goat_errors}) ($i$) \emph{partial matches}, where the instruction fits the target only in part; ($ii$) \emph{hallucinations}, where attributes mentioned in the text are absent from the scene; and ($iii$) \emph{mesh artefacts}, where incomplete geometry in HM3D-Sem hides the object, rendering the description misleading;  ($iv$) \emph{reference to object bounding box}, where the goal descriptions contain references to object bounding box, which the agent might not have access to during navigation (e.g. `...region defined by the bathtub bounding box'); ($v$) \emph{spatial errors}, where the spatial relationship is wrong between objects (e.g. `blanket near the bed on the left side' when it is clearly towards the foot of the bed in the figure). 
To obtain a statistic on how often these five error types occur, we randomly sample 100 goals from GOAT-Bench and manually inspect the descriptions along with the corresponding goal images. We found that only 33\% of them were accurate whereas the rest had errors - 22\% were hallucination errors, 7\% were mesh artefact errors, 3\% were partial match errors, 23\% were spatial errors and 12\% were due to reference to object bounding box or the image frame, totaling to 67\% error. The most frequent error types were spatial errors and hallucination errors.
Moreover, in GOAT-Bench the language goals are generated to admit a single matching target, and they were not systematically validated to rule out additional scene objects that satisfy the same description.

\begin{figure}[t]
  \centering
  \includegraphics[width=\linewidth]{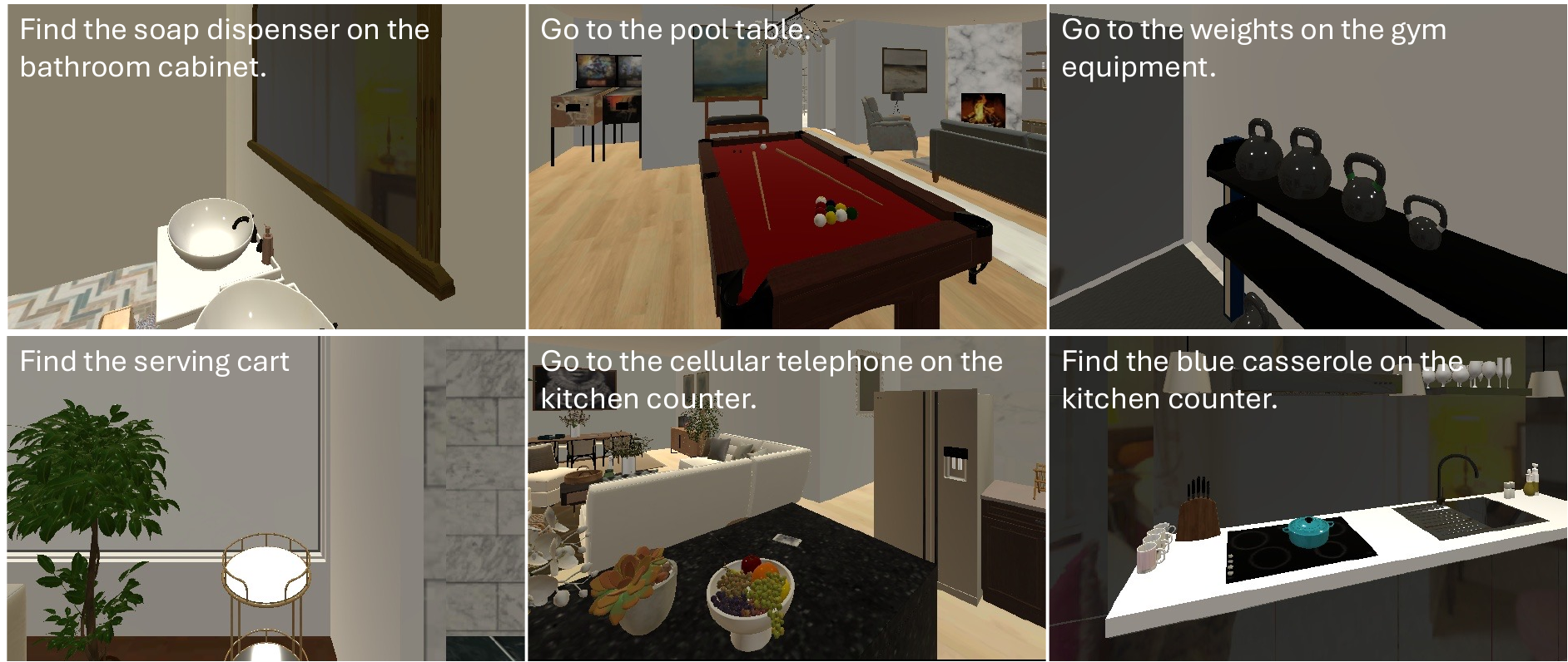}
  \caption{In \DATASET, we use objects and attributes available in HSSD synthetic scenes, thus producing error free language descriptions.
  }
  \label{fig:dataset_visuals}
\end{figure}

We mitigated these errors by first using scenes and objects from HSSD dataset. The scenes are synthetic and hence free of mesh-related issues. The dataset also provides ground-truth categories and attributes for every object; we pass this structured metadata to GPT to obtain fluent natural-language instructions (\Cref{sec:supp_llm_usage}) that remain anchored to the ground truth. All generated sentences are then manually reviewed and, where necessary, corrected to eliminate any residual LLM errors (see \Cref{fig:dataset_visuals}). Finally, we tag each instruction with the linguistic cues used in our benchmark (colour, size, material, state, and so on) (\Cref{sec:supp_llm_usage}) and include these annotations with the dataset—information that no previous semantic-navigation corpus offers.
Moreover, all possible correct goal locations corresponding to a single description are added in the episode to account for multiple matches. 

\subsubsection{Viewpoint generation}
\label{sec:viewpoint_gen}
Similar to the ObjectNav task \citep{batra2020objectnav}, we store \emph{viewpoints} or navigable positions around the goal object, such that navigating near to any of those will be considered successful. To generate candidate viewpoints around an object, we use depth observations together with the object’s 3-D position and dimensions to obtain viewpoints with a clear line of sight to the target object. First, we trace the object’s non-navigable boundary by sweeping a full $360^o$ and recording every transition point from free space to obstacle.
Around each boundary point we then place radial samples at uniform intervals out to 1.5 m.
Each sample is accepted as a valid viewpoint only if it meets two criteria: the location is navigable, and—after orienting the agent toward the object—no intervening obstacle blocks the view. We describe the full algorithm below.

\newcommand{\tts}[1]{\texttt{\small #1}\xspace}
\newcommand{\vobjpos}{\tts{obj\_pos}\xspace}
\newcommand{\vobjdims}{\tts{obj\_dims}\xspace}
\newcommand{\vdistbound}{\ensuremath{\tts{dist}_\texttt{bound}}\xspace}
\newcommand{\vradiusbound}{\ensuremath{\tts{radius}_\texttt{bound}}\xspace}
\newcommand{\vdistvp}{\ensuremath{\tts{dist}_\texttt{vp}}\xspace}
\newcommand{\vradiusvp}{\ensuremath{\tts{radius}_\texttt{vp}}\xspace}
\newcommand{\vboundarypts}{\ensuremath{\tts{boundary\_points}}\xspace}
\newcommand{\vptbound}{\ensuremath{\tts{pt}_\texttt{bound}}\xspace}
\newcommand{\vradialpts}{\tts{radial\_pts}\xspace}
\newcommand{\vptradial}{\ensuremath{\tts{pt}_\texttt{radial}}\xspace}
\newcommand{\vviewpts}{\tts{view\_pts}\xspace}
\newcommand{\vpixelobjdims}{\ensuremath{\tts{pixel}_\texttt{obj\_dims}}\xspace}

\begin{algorithm}[H]
\caption{generate\_view\_points}
\begin{algorithmic}[1]
\REQUIRE Input data \vobjpos, \vobjdims, \vdistbound, \vradiusbound, \vdistvp, \vradiusvp 
\ENSURE Output \textit{Viewpoints around the object}

\STATE Obtain \vboundarypts around the object
\FOR{$\theta$ in $(0, 2\pi)$}
    \STATE Next equidistant \vptbound $\leftarrow$ Binary\_Search (\vdistbound, \vradiusbound, $\theta$ ) 
    \IF{\vptbound  exists}
        \STATE Append \vptbound to \vboundarypts
    \ENDIF
\ENDFOR

\STATE \vviewpts $\leftarrow \{\}$ 
\FOR{\vptbound in \vboundarypts}
\STATE Compute equidistant \vradialpts $\leftarrow$ $f(\vradiusvp, \vdistvp)$
\FOR{\vptradial in \vradialpts}
    \STATE Skip if within previous \vptbound area
    \STATE Ensure \vptradial is navigable
    \STATE Face the Agent towards \vobjpos $\leftarrow$ $f(\vptradial, \vobjpos)$
    \STATE Compute corresponding pixel locations of \vobjdims
    \STATE From \textit{Depth} observation, obtain depth values at above pixel values
    \IF{depth at \vpixelobjdims $>$ distance to \vobjdims }
        \STATE Append to \vviewpts
    \ENDIF
\ENDFOR
\ENDFOR
\RETURN \vviewpts

\end{algorithmic}
\end{algorithm}

\subsection{\methodnameshort method}
\label{supp:method-more}
\subsubsection{\egoimagemap baseline}
\label{supp:llm-egoimage}
For the \egoimagemap baseline method, we prompt a GPT-5 model with the following prompt to infer spatial relations between two objects in an egocentric RGB image:

\begin{quote}
``\texttt{You are a vision-language assistant. You are given a query and a group of images. Each group contains several images. For every image in each group, return whether the object in the query is clearly visible.} 

\texttt{Respond strictly in JSON. Group numbers are integers (0, 1, 2, …). Use lowercase booleans (`true` / `false`). Do not include explanations, only the JSON.}''
\end{quote}

\subsubsection{\mlfmvlm}
\label{supp:llm-mlfmvlm}
In our \mlfmvlm variant, we prompt GPT-5 to infer the most likely relation between two objects from a list of images:

\begin{quote}
``\texttt{Your task is to infer the most likely relation between object A (colored in red) and object B (colored in blue) from the list of images.}

\texttt{Note that the images are top-down map projections and image 1 is physically at a lower height than image 2 which is lower than image 3 and so on. In these images, you will find a green box, denoting the agent location. Return a json with "relation" and "rationale" fields.}

\texttt{Follow these rules. Use Euclidean proximity for near/next to (next to = very close, non-overlapping; near = close but not touching). If green box for the agent is present, decide LEFT/RIGHT relative to the ray from agent toward the midpoint between A and B. Consider the nearest blobs for A and B to the agent. Be concise and avoid extra prose.}''
\end{quote}

\subsubsection{\mlfmrgraph} 
\label{sec:rgraph_more}
In this section, we provide more detail into how we construct the \mlfmrgraph.
The agent incrementally builds and updates the relation graph during navigation. As it encounters new views, nodes may be added or merged, and edges updated if new relational evidence is observed. Importantly, since objects may extend over multiple adjacent grid cells in the map, we collapse these into a single graph node and record the set of grid cells spanned by the object ($map\_span$). This aggregation serves two purposes: (1) it yields a segmentation-like mask of the object's spatial extent, and (2) it allows more accurate reasoning about its relation to nearby objects, which can now be computed relative to the object's full footprint rather than a single cell.

To infer relations between two nodes, we apply geometric rules on their $map\_span$, which records the grid cells occupied by each object as $[x,y,z]$ coordinates ($(x,y)$ for 2D grid location and $z$ for map layer). Vertical relations (above, below) are determined by comparing $z$ values, while proximity is inferred if objects are within $\sim$20 grid cells (grid resolution is 6 cm). Using the agent's pose, we transform object locations into the egocentric frame and their horizontal displacement then yields left or right relations. Containment is inferred by checking whether A's boundaries lie mostly within B's boundaries. We record all relations satisfying these rules for each node pair and store their corresponding text features in the edge. These geometric rules serve as a deterministic first step for relation inference; these can be replaced with a learned model trained on collected episodes to resolve ambiguous cases and support unseen relation types.

During the exploitation phase, the agent uses the RGraph to answer relational queries. Given a text query, we encode it using the CLIP text encoder and compute its cosine similarity with the stored graph features. Specifically, for each edge, we get (a) visual features for the source node, (b) visual features for the sink node and (c) the edge text features. The agent selects the candidate target node whose triplet score (target-relation-reference) is highest and exceeds a confidence threshold.

This formulation allows the agent to reason jointly over object semantics and spatial relations captured in the RGraph. Moreover, comparing CLIP text embeddings of the relational word allows the model to generalize to unseen but semantically similar relations (e.g., `under' and `below', `within' and `inside', `beside' and `next to'). In effect, RGraph provides a structured interpretable memory that grows with exploration and enables reasoning over spatial relations during exploitation.

\subsection{Semantic navigation baselines}
\label{sec:baselines_more}
In this section, we briefly describe the baselines against which we compare our method in the main paper.

\mypara{VLMaps}~\citep{huang2023visual} store LSeg~\citep{li2022language} pixel-level embeddings in a 2D top-down grid map, along with their 3D location. They average the pixel embeddings when multiple pixels are projected onto the same grid and also across multiple views.

\mypara{VLFM}~\citep{yokoyama2023vlfm} builds a 2D value map that stores BLIP-2~\citep{li2023blip} similarity scores of the observed images and object category, and uses it to semantically explore the environment. It achieves SOTA performance in the ObjectNav task. 

\mypara{MOPA}~\citep{Raychaudhuri_2024_WACV} iteratively builds a 2D category map after detecting objects in the frame, randomly explores the environment and navigates to the goal once found. We replace their PointNav~\citep{wijmans2019dd} with the A* planner for fair comparison.

\mypara{OneMap}~\citep{busch2024mapallrealtimeopenvocabulary} builds a 2D feature map with SED~\citep{xie2024sed} patch features by using a confidence-based fusion mechanism. 
They employ an open-set object detector YOLO-World~\citep{cheng2024yolo} along with the map to identify the target and use an A* planner to navigate.

\subsection{Ablations}
\label{sec:ablations_more}
In this section, we report ablations to determine how the different components contribute to \methodnameshort's performance. We ablate on grid resolution in \Cref{subsec:grid_res_ablation}, followed by EAE percentage in \Cref{subsec:eae_ablation}, and the contribution of map and object detector in \Cref{subsec:map_vs_detector_ablation}. 

\subsubsection{Effect of grid resolution and number of layers}
\label{subsec:grid_res_ablation}

\begin{table}[t]
\caption{
     Varying map resolution and number of layers in our multi-layered map contributes towards \methodnameshort's performance.
     }  
\label{tab:ablation_num_layers}
\centering
\begin{tabular}{p{0.1cm}ccrr}
\toprule
&cm per 2D grid cell &Number of layers &SR$\uparrow$ &SPL$\uparrow$ \\
% \cmidrule{4-5}
\midrule

1) &6 & 1 &29.7 &9.3 \\

2) & & 2 &37.0 &14.1 \\

3) & & 3  &\textbf{39.5} & \textbf{14.9}  \\

4) & & 4 &38.5 & 13.8 \\

5) & &5 & 27.6 &10.1  \\

% \cmidrule{2-13}
\midrule

6) & 10 &3 &25.1 &9.6  \\

\bottomrule
\end{tabular}
\end{table}

\Cref{tab:ablation_num_layers} reports an ablation on the two key map hyper-parameters: the number of height layers \(L\) and the \((x,y)\) resolution of each slice. Success rises as \(L\) grows from one to three layers, but degrades when a fourth layer is added. Beyond three slices large objects become fragmented, making them harder to localise; we therefore set \(L=3\) in all main experiments.
Keeping \(L\) fixed, performance improves when we refine the \((x,y)\)-grid—i.e.\ when the centimetres-per-cell value decreases (rows 1 vs.\ 6).  
A finer resolution allows the map to preserve small objects' geometry and thus boosts the agent’s success rate.

\subsubsection{Effect of explore-and-exploit (EAE) percentage}
\label{subsec:eae_ablation}
%We ablate on the EAE percentage in \Cref{tab:ablation_eae} to understand its contribution in the agent performance. We notice that the overall SR and SPL are the best when we do 40\% of explore-and-exploit (EAE) with 60\% exploit-only (E). 
%We also report the error percentage for each row in \Cref{tab:ablation_eae}. It shows that the `wrong goal on map' goes down when we explore more, indicating that the built map becomes more and more accurate on exploring more. However, exploring more leads to more out-of-time errors since the agent is left with less time to exploit.
\Cref{tab:ablation_eae} varies the share of the episode spent to Explore{\nobreakdash-and\nobreakdash-Exploit} (EAE) versus Exploit-only~(E). Overall performance peaks when the agent spends the first $40\%$ of its time budget in EAE and the remaining $60\%$ in E: both \emph{Success Rate} ($SR$) and \emph{Success weighted by Path Length} ($SPL$) achieve their highest values under this $40\!:\!60$ split.
The table also reports the error composition. As the EAE portion grows, the \emph{wrong-goal-on-map} rate declines,  indicating that extended exploration yields a more accurate map. However, longer exploration leaves less budget for exploitation, so \emph{out-of-time} failures increase once the EAE share exceeds $40\%$.

\begin{table}[t]
\caption{
     The agent achieves the best performance when it spends 40\% in Explore-and-Exploit (EAE) and the remaining time in Exploit-only phase.
     }  
     
\label{tab:ablation_eae}
\centering
\begin{tabular}{lcccccc}
\toprule
\multirow{2}{*}{\thead{EAE\\(\%)}} &\multirow{2}{*}{SR$\uparrow$} &\multirow{2}{*}{SPL$\uparrow$} &\multicolumn{3}{c}{Error$\downarrow$}  \\

&& & & Wrong detection &Wrong goal on map & OOT \\
% \midrule
\cmidrule{2-3} \cmidrule{5-7}

20 &36.7 &11.9  &&\textbf{25.1} & 26.0 & \textbf{48.9} \\

40 &\textbf{39.5} & \textbf{14.9} & &31.7 &17.1 & 51.2 \\

60 &36.6 &9.8 & &28.7 & 15.5 & 55.8 \\

80 &33.3 &9.6  & &28.5 &\textbf{8.2} &63.3 \\

\bottomrule
\end{tabular}
\end{table}

% \begin{table}[ht]
% \caption{
%      Performance as we vary the percentage of time spent in Explore-and-Exploit (EAE).  The remaining time is spent in Exploit-only.
%      }  
     
% \label{tab:ablation_eae}
% \centering
% \resizebox{\linewidth}{!}{
% \begin{tabular}{lcc|ccccccccc|ccc}
% \toprule
% \multirow{2}{*}{\thead{EAE\\(\%)}} &\multirow{2}{*}{SR$\uparrow$} &\multirow{2}{*}{SPL$\uparrow$} &\multicolumn{9}{c}{SR$\uparrow$} &\multicolumn{3}{c}{Error$\downarrow$}  \\

% && &no &color &size &texture &number  &material &state &modifier &support & \thead{Wrong\\detection} &\thead{Wrong goal\\on map} & OOT \\
% \midrule

% 20 &\textbf{43.6} &16.3 &37.8 &\textbf{38.5} &41.0 &0.0 &\textbf{14.3} &59.0 &\textbf{14.3} &\textbf{27.9} &27.0 &\textbf{25.1} & 26.0 & \textbf{48.9} \\

% 40 &\textbf{43.6} & \textbf{16.9} &\textbf{37.9} &\textbf{38.5} &\textbf{41.7} &0.0 &\textbf{14.3} &\textbf{64.4} &\textbf{14.3} &\textbf{27.9} &\textbf{27.3} &31.7 &17.1 & 51.2 \\
% 60 &43.0 &15.1 &13.9 &32.9 &39.2 &0.0 &14.3 &40.3 &14.3 &20.3 &27.3 &28.7 & 15.5 & 55.8 \\

% 80 &39.7 &11.6 &13.6 &29.4 &31.1 &0.0 &0.0 &31.7 &0.0 &17.7 &22.0 &28.5 &\textbf{8.2} &63.3 \\

% \bottomrule

% \end{tabular}
% }
% \end{table}

\subsubsection{Role of map and object detector in \methodnameshort}
\label{subsec:map_vs_detector_ablation}

\begin{table}[t]
\caption{
     \methodnameshort achieves the best result when both the map and the object detector are used.
     }  
     
\label{tab:ablation_map}
\centering
\begin{tabular}{cccc}
\toprule
Map &Object detector &SR$\uparrow$ &SPL$\uparrow$  \\

\midrule

$\checkmark$ & $\times$ & 32.8 &14.1 \\

$\times$ & $\checkmark$ &29.0 & 8.6 \\

$\checkmark$ & $\checkmark$ &\textbf{39.5} & \textbf{14.9}  \\

\bottomrule
\end{tabular}
\end{table}

% \begin{table}[ht]
% \caption{
%      \textbf{Role of map and object detector} in \methodnameshort.
%      }  
     
% \label{tab:ablation_map}
% \centering
% \resizebox{\linewidth}{!}{
% \begin{tabular}{cccc|ccccccccc}
% \toprule
% \multirow{2}{*}{Map} &\multirow{2}{*}{Object detector} &\multirow{2}{*}{SR$\uparrow$} &\multirow{2}{*}{SPL$\uparrow$} &\multicolumn{9}{c}{SR$\uparrow$} \\

% &&& &no &color &size &texture &number  &material &state &modifier &support \\

% \midrule

% $\checkmark$ & $\times$ & 43.0 &16.1 & \textbf{42.7} &30.7 &\textbf{41.7} &0.0 &\textbf{21.4} &61.1 &14.3 &\textbf{28.2} &27.3 \\

% $\times$ & $\checkmark$ &31.1 & 10.9 & 16.7 &9.7 &32.3 &0.0 & 7.1 &43.7 &0.0 & 18.9 &13.4  \\

% $\checkmark$ & $\checkmark$ &\textbf{43.6} & \textbf{16.9} &37.9 &\textbf{38.5} &\textbf{41.7} &0.0 &14.3 &\textbf{64.4} &14.3 &27.9 &27.3 \\

% \bottomrule

% \end{tabular}
% }
% \end{table}

%In this section, we try to understand the role that the object detector and the multi-layer map play in the agent performance. \Cref{tab:ablation_map} shows that the map-only version (row 1) performs similar to the one that uses both the map and the detector (row 3) on the overall SR and SPL. This indicates that our multi-layer map representation is extremely effective. However the detector only version (row 2) performs the worst among all the versions.
We next assess the individual contributions of the object detector and the multi-layer map. \Cref{tab:ablation_map} compares three variants: map~only, detector~only, and the full system that combines both.
Using both achieves the best performance followed by the \emph{map-only} version followed by \emph{detector-only} version. 
% The \emph{map-only} agent (row~1) matches the full model (row~3) on both $SR$ and $SPL$, confirming that the multi-layer representation by itself is highly effective.  
% By contrast, the \emph{detector-only} variant (row~2) lags well behind, highlighting the importance of the spatial memory supplied by the map.

% \subsubsection{Role of text-as-a-kernel querying technique}
% \input{tables/tab-ablation-querying}
% \label{subsec:text_as_kernel_ablation}
% We compare how the agent performance is affected when we query our multi-layer map without and with the text-as-a-kernel approach. It is evident from \Cref{tab:ablation_querying} that the text-as-a-kernel improves the support relationships by a large margin, thus leading to better overall performance.

% \input{sec/experiments-suppl}

\section{Possible future work}
\label{sec:limitations_more}

The language in \DATASET\ can be improved to include complex linguistic structures—coreference (e.g., “place \emph{it} on the table”) and negation (“a chair that is \emph{not} black”) and action directives beyond navigation verbs (e.g. “pick up”, “bring back”).
Our dataset can be further augmented with additional kinds of spatial relations such as egocentric proximity (e.g. ``the chair near you''),  allocentric relations based on object placement (e.g. ``the chair in the corner of the room'') or relational with respect other instances of the same objects (e.g. ``the taller plant''). We leave this for future work.
% We also acknowledge that although one could define a third type of reference frame beyond egocentric and allocentric - the ``world'' frame - which specifies relationships in a fixed global coordinate system (e.g., ``go to the couch five meters to the north''), we deliberately exclude this category, as such formulations are not linguistically natural and are uncommon in human instructions.
Our dataset generation approach makes a few assumptions. Bounding boxes only approximate true object geometry, which can lead to mislabels for irregularly shaped objects (e.g., a table placed in front of an L-shaped sofa may be incorrectly labeled as ``inside'' the sofa because its bounding box lies within that of the sofa). Fixed proximity thresholds (0.2 meter) may not generalize across scales (e.g., ``near'' may imply a few centimeters for tabletop objects but several meters in an office setting when describing ``the water cooler near the meeting room''). Future work could incorporate more perceptually grounded measures, such as learned priors for spatial relations or scene-normalized distance metrics (e.g., defining ``near'' as within 10\% of a room's longest dimension).
% Finally, there is significant room for improvement in agent performance. In this work we highlight how various querying techniques can be used to effectively query the multi-layer feature map and discuss their limitations. Future work could combine VLM and relation graph or use a learned model to further improve the agent's spatial reasoning.}
% Extending the dataset with these linguistic phenomena would broaden its coverage and support more advanced studies of language understanding in robotics and embodied AI.

\end{document}